\documentclass[lettersize,journal]{IEEEtran}
\usepackage{amsmath,amsfonts}
\usepackage{algorithmic}
\usepackage{algorithm}
\usepackage{array}
\usepackage[caption=false,font=normalsize,labelfont=sf,textfont=sf]{subfig}
\usepackage{textcomp}
\usepackage{stfloats}
\usepackage{url}
\usepackage{verbatim}
\usepackage{graphicx}
\usepackage{cite}
\usepackage{booktabs}
\usepackage[hidelinks]{hyperref}
\usepackage{multirow}
\usepackage{multicol}

\begin{document}

\title{Towards Robust Adaptation of Graph Foundation Models under Label Noise%
\thanks{\textcopyright\ 2026 IEEE. Personal use of this material is permitted.
Permission from IEEE must be obtained for all other uses, in any current or
future media, including reprinting/republishing this material for advertising
or promotional purposes, creating new collective works, for resale or
redistribution to servers or lists, or reuse of any copyrighted component
of this work in other works.}}

\author{Rui Zhao, Bin Shi,~\IEEEmembership{Member,~IEEE,} and Bo Dong,~\IEEEmembership{Member,~IEEE}
\thanks{
Rui Zhao and Bin Shi are with the School of Computer Science and Technology, Xi'an Jiaotong University, and also with the Shaanxi Province Key Laboratory of Big Data Knowledge Engineering, Xi'an 710049, China. (e-mail: rayn\_z@stu.xjtu.edu.cn; shibin@xjtu.edu.cn)}
\thanks{Bo Dong is with the School of Distance Education,
Xi'an Jiaotong University, Xi'an 710049, China. (e-mail: dong.bo@xjtu.edu.cn)}%
}



% \IEEEpubid{© 2026 IEEE. Personal use of this material is permitted. Permission from IEEE must be obtained for all other uses, in any current or future media, including reprinting/republishing this material for advertising or promotional purposes, creating new collective works, for resale or redistribution to servers or lists, or reuse of any copyrighted component of this work in other works.}

\maketitle

\begin{abstract}
Graph foundation models (GFMs) have shown promising transferability across tasks, but adapting them to downstream settings still relies heavily on fine-tuning. Under label noise, however, adaptation becomes particularly challenging: GFMs cannot reliably detect corrupted labels before adaptation, yet direct supervised fine-tuning on noisy data may distort the useful priors learned during pre-training. To address this issue, we propose Prototype-calibrated Adaptation with Collaborative Tuning (PACT), a teacher-student distillation framework in which a robust teacher distills guidance to the student, aided by prototypes derived from pre-trained GFM priors. Specifically, PACT first employs a parameter-efficient LoRA teacher, which is less prone to memorizing noisy labels, to provide stable supervision for a high-capacity student and thus prevent pre-trained GFM representations from being rapidly distorted during full fine-tuning. With these representations preserved, PACT further introduces a graph-aware prototype calibration module that constructs class prototypes from frozen representations for reliable sample selection and label correction. Extensive experiments on multiple benchmark datasets across diverse noise settings demonstrate the effectiveness and robustness of PACT.
\end{abstract}


\section{Introduction} \label{sec:intro}
In recent years, \textit{Graph Foundation Models} (GFMs) have emerged as a transformative paradigm in graph machine learning, leveraging massive-scale pre-training to capture universal structural and semantic priors \cite{wang2025graph}. Despite their strong representational capacity, the zero-shot performance of GFMs on downstream tasks is often suboptimal due to significant distribution shifts in both feature spaces and structural patterns \cite{liu2023towards}. Consequently, adapting these pre-trained models to specific domains through fine-tuning has become an indispensable step to bridge the gap between generalized knowledge and specialized downstream requirements \cite{hu2019strategies,liu2025graph}.

Recent studies have proposed various adaptation strategies to better align GFMs with downstream tasks \cite{liu2024one,zhao2025graphany,sun2025riemanngfm}. However, most of these methods assume access to clean annotations. In real-world graph mining scenarios, clean annotations are often difficult to guarantee \cite{dai2021nrgnn}, leaving existing adaptation methods vulnerable to label noise.

A straightforward solution is to directly apply existing \textit{noisy label learning} (NLL) methods during fine-tuning. However, most NLL methods require an initial classifier to provide the signals used for noise handling, such as prediction confidence \cite{zhao2024training}, training loss \cite{jiang2018mentornet,li2020dividemix}, or consistency signals \cite{wei2020combating}. In practice, such a classifier is usually obtained either from a small warm-up model trained under noisy supervision \cite{patrini2017making,dai2021nrgnn,li2025leveraging} or from a general-purpose zero-shot model \cite{wang2024noisegpt,pan2025nlprompt,zhang2025lemon} (e.g., CLIP \cite{radford2021learning} or MMICL\cite{zhao2023mmicl}), while neither option is well suited to GFMs. Specifically, most GFMs do not come with a readily usable zero-shot classifier for downstream tasks, while a noisy warm-up process may already distort the valuable knowledge encoded in the pre-trained model before robust adaptation begins.

\begin{figure}[t]
    \centering
    \includegraphics[width=0.95\linewidth]{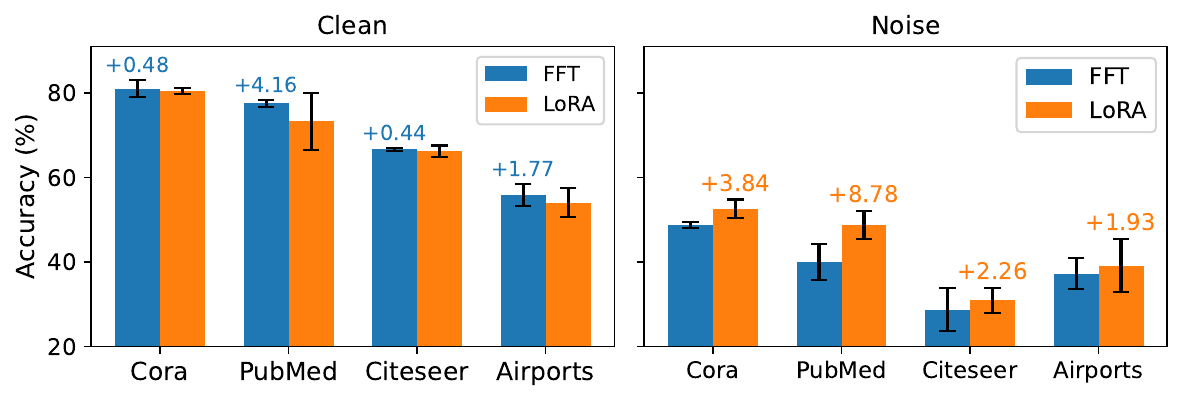}
    \caption{Comparison between PEFT (LoRA) and FFT under clean (left) and uniform 50\% noisy (right) adaptation conditions. Here, results are obtained by adapting RiemannGFM \cite{sun2025riemanngfm} under its standard setting, comparing LoRA applied to the linear layers with direct full fine-tuning. Error bars denote standard deviations over 5 runs.}
    \label{fig:lora_fft_intro}
\end{figure}

\IEEEpubidadjcol
These limitations indicate that the main challenge lies not only in noise handling itself, but also in how to adapt GFMs without destroying their pre-trained priors. This motivates us to examine how different adaptation strategies behave under noisy supervision. As illustrated in Figure \ref{fig:lora_fft_intro}, we compare \textit{full fine-tuning} (FFT) with \textit{parameter-efficient fine-tuning} (PEFT, instantiated by LoRA \cite{hu2022lora}). The results show that FFT performs better than LoRA in the clean setting, but experiences a larger performance drop under noisy adaptation. This contrast reveals a useful complementarity between the two adaptation regimes: FFT offers stronger adaptation capacity, while PEFT provides better robustness to noisy supervision \cite{wei2024vision,kim2024towards}.

To harness these complementary strengths, we introduce Prototype-calibrated Adaptation with Collaborative Tuning (PACT), a novel framework that couples a stable PEFT teacher with a high-capacity FFT student. Specifically, PACT is designed around two synergistic goals: suppressing noise memorization during adaptation while exploiting pre-trained priors to mitigate label noise. To achieve the first goal, we design a LoRA-FFT teacher-student architecture. Operating within a restricted parameter space, the LoRA teacher acts as a robust filter that screens the corrupted supervision, thereby providing the FFT student with purified learning signals to avoid direct noise corruption. To fulfill the second goal, PACT introduces a graph-aware prototype calibration mechanism that leverages frozen pre-trained representations, which remain insulated from the noisy adaptation process, to help construct robust class prototypes. These prototypes are then used to recalibrate noisy labels, thereby reducing the risk of error accumulation. Finally, guided by these calibrated signals, the FFT student is optimized for more effective downstream adaptation under noisy supervision.

Our main contributions can be summarized as follows:
\begin{itemize}
    \item We study the problem of adapting Graph Foundation Models (GFMs) under label noise, where noisy warm-up can easily damage pre-trained GFM priors, making existing noisy label learning methods less suitable for this setting.
    
    \item We introduce Prototype-calibrated Adaptation with Collaborative Tuning (PACT). PACT uses a robust LoRA teacher to filter noisy supervision for a high-capacity FFT student, while leveraging frozen GFM representations decoupled from noisy adaptation for prototype-based label recalibration, thereby enabling more robust and effective GFM adaptation under label noise.
    
    \item Extensive experiments on multiple benchmark datasets under diverse label noise settings demonstrate that PACT outperforms baselines and confirm the effectiveness of both collaborative tuning and prototype calibration.
\end{itemize}

\section{Related Work}
In this section, we review prior studies most relevant to PACT from two perspectives: graph foundation model adaptation and graph learning under label noise. While these two directions provide important foundations for our method, their combination has received limited attention. The following discussion positions PACT in relation to these lines of work.
\subsection{Graph Foundation Models and Adaptation Strategies}
Graph foundation models have significantly advanced graph machine learning by pre-training on large-scale datasets to learn transferable structural and semantic priors \cite{bommasani2021opportunities,sun2026survey,xu2026generalist}. Recent works, such as GraphAny \cite{zhao2024graphany}, PRODIGY \cite{huang2023prodigy}, OFA \cite{liu2024one}, and RiemannGFM \cite{sun2025riemanngfm}, aim to unify diverse graph data into a shared representation space, allowing the models to generalize across different domains. However, inherent discrepancies in feature distributions, graph structures, and task objectives between pre-training and downstream settings often limit the effectiveness of direct transfer \cite{hu2019strategies,liu2023towards,sun2025riemanngfm}. As a result, downstream adaptation remains essential for fully exploiting the priors encoded in pre-trained GFMs \cite{wang2025graph}.

To bridge this gap, various downstream adaptation strategies have been explored. A conventional approach is \textit{full fine-tuning} (FFT) \cite{hu2019strategies}, which updates all pre-trained parameters for task-specific adaptation. To improve efficiency and better preserve pre-trained knowledge, recent studies have increasingly turned to \textit{parameter-efficient fine-tuning} (PEFT) \cite{cheng2026prompt,zhang2025vpt}. PEFT methods generally branch into two directions: (1) architecture-based tuning, which inserts or updates a minimal set of trainable parameters, such as adapters or low-rank modules \cite{li2024adaptergnn,yang2025graphlora}; and (2) prompt-based tuning, which keeps the model backbone strictly frozen and bridges the downstream task gap using learnable graph prompts \cite{sun2022gppt,yu2025non,yu2025gcot,sun2023graph,huang2025one}. Despite the rapid progress in graph adaptation, robustness under noisy supervision remains largely underexplored. Existing studies mainly focus on efficient transfer under standard downstream settings, while recent findings in vision and language suggest that the restricted parameter space of PEFT may naturally hinder rapid memorization of noisy labels \cite{wei2024vision}. Motivated by this observation, we explore whether such a property can be leveraged for noisy graph foundation model adaptation through collaborative tuning and prototype-based calibration.

\subsection{Learning on Graphs under Label Noise}
Learning with noisy labels has been extensively studied to improve model robustness. In the graph domain, this problem is particularly challenging because standard \textit{Graph Neural Networks} (GNNs) are highly vulnerable to label noise: their message-passing mechanism can propagate, and even exacerbate, incorrect supervision from mislabeled nodes to their neighbors \cite{dai2021nrgnn,wang2024noisygl}.

Existing research on graph noisy-label learning can be categorized into two types. 
\textit{Label-centric} methods mainly improve supervision quality by identifying unreliable annotations and refining them through graph-specific operations such as pseudo-label correction \cite{ding2024divide}, label propagation\cite{xia2023gnn}, or reliable-sample selection\cite{qian2023robust}. 
In contrast, \textit{representation-centric} methods aim to learn noise-robust representations \cite{cheng2026isolating}, for example through structure enhancement \cite{dai2021nrgnn} or consistency regularization \cite{li2025leveraging}, so as to mitigate the negative impact of corrupted labels during message passing \cite{wang2024noisygl}. Despite this progress, existing methods are primarily developed for task-specific GNNs trained directly on downstream graphs. As a result, they mainly study how to suppress noise fitting or improve label reliability during graph training, rather than how to perform robust adaptation when the model already carries valuable pre-trained priors, as in GFMs. This leaves open a different and important question: how to exploit the structural priors already encoded by pre-trained GFMs for noise reduction, without simultaneously corrupting them during downstream adaptation.

\section{Preliminaries and Problem Formulation}
This section introduces the background and notation used throughout the paper. It starts with the preliminaries of graph foundation model adaptation, including the instantiation of pre-trained GFMs for downstream tasks and common adaptation strategies, and then moves to the formulation of graph adaptation under label noise.
\subsection{Preliminaries} \label{sec:pre}
\smallskip\noindent\textbf{Graph foundation models.} In contrast to traditional GNNs trained from scratch on a single dataset, GFMs are pre-trained on large-scale and diverse graph corpora to capture transferable structural and semantic priors. The goal is to learn a general-purpose graph encoder $f_{\theta}$, parameterized by $\theta$, that captures reusable knowledge across different graph domains. Formally, given an attributed graph $\mathcal{G}$ with adjacency matrix $A$ and node feature matrix $X$, a GFM encodes the input graph into transferable latent representations:
\begin{equation}
    H = f_{\theta}(A, X),
\end{equation}
where $H$ denotes the representations produced by the pre-trained encoder. These representations can be further adapted to downstream tasks through task-specific heads or parameter updating strategies.

\smallskip\noindent\textbf{Adaptation of GFMs.} In this work, we focus on adapting pre-trained GFMs to downstream supervised classification tasks. Inspired by recent unified graph learning paradigms \cite{liu2024one}, we view downstream GFM adaptation as learning classification predictions over a set of target instances $\mathcal{S}=\{s_i\}_{i=1}^{N}$. Depending on the task granularity, an instance $s_i$ may correspond to a node, a node pair, or a graph. The downstream supervision is given by a labeled dataset $\mathcal{D}=\{(s_i,y_i)\}_{i=1}^{N}$, where $y_i \in \mathcal{Y}=\{1,2,\dots,C\}$ and $C$ is the number of classes. Given a pre-trained graph foundation model encoder $f_{\theta}$ and a task-specific classifier head $g_{\phi}$ with parameter $\phi$, the adapted predictor is defined as:
\begin{equation}
    h_{\psi}(s_i)=g_{\phi}\big(f_{\theta}(A_i, X_i)\big),
\end{equation}
where $(A_i, X_i)$ denotes the adjacency matrix and feature matrix of the contextual graph view associated with instance $s_i$, and $\psi=\{\theta,\phi\}$ denotes the parameters of the predictor. The goal of downstream adaptation is to learn a predictor $h_{\psi}: \mathcal{S} \rightarrow \mathcal{Y}$ from $\mathcal{D}$ that maps each target instance $s_i$ to its label $y_i$.

\smallskip\noindent\textbf{Adaptation strategies: FFT and PEFT.} To optimize the predictor $h_{\psi}$, two standard adaptation paradigms are widely used. FFT updates the entire parameter set $\psi=\{\theta,\phi\}$ on downstream data, with the objective:
\begin{equation}
    \min_{\theta,\phi}\frac{1}{N}\sum_{i=1}^{N}\ell\big(h_{\psi}(s_i),y_i\big),
\end{equation}
where $\ell$ denotes the classification loss. In contrast, PEFT updates only a small subset of task-specific parameters while keeping most pre-trained parameters fixed. Its objective can be written in a general form as:
\begin{equation}
    \min_{\Delta\psi}\frac{1}{N}\sum_{i=1}^{N}\ell\big(h_{(\psi^{*},\Delta\psi)}(s_i),y_i\big),
\end{equation}
where $\psi^{*}$ denotes the subset of pre-trained parameters kept frozen during adaptation, and $\Delta\psi = \{\Delta\theta, \phi\}$ represents the trainable parameter set, which includes the injected parameter-efficient modules $\Delta\theta$ and the task-specific head $\phi$. Therefore, FFT and PEFT reflect two complementary adaptation regimes: FFT favors stronger adaptation capacity, whereas PEFT better preserves pre-trained knowledge and can be more robust under noisy supervision.

\subsection{Problem Formulation}
Let $\mathcal{D}=\{(s_i, y_i)\}_{i=1}^{N}$ denote the clean downstream dataset, where $s_i$ is a target instance and $y_i \in \mathcal{Y}$ is its ground-truth label. In this work, we study graph foundation model (GFM) adaptation under label noise, where the learner only observes the noisy dataset $\widetilde{\mathcal{D}}=\{(s_i, \tilde{y}_i)\}_{i=1}^{N}$, and the annotated label $\tilde{y}_i$ may differ from $y_i$. Formally, the goal of adaptation under label noise is to learn an adapted predictor $h_{\psi}: \mathcal{S} \rightarrow \mathcal{Y}$ from the noisy dataset $\widetilde{\mathcal{D}}$ that correctly maps each target instance $s_i$ to its underlying ground-truth label $y_i$.

\begin{figure*}[!t]
\centering
\includegraphics[width=0.95\textwidth]{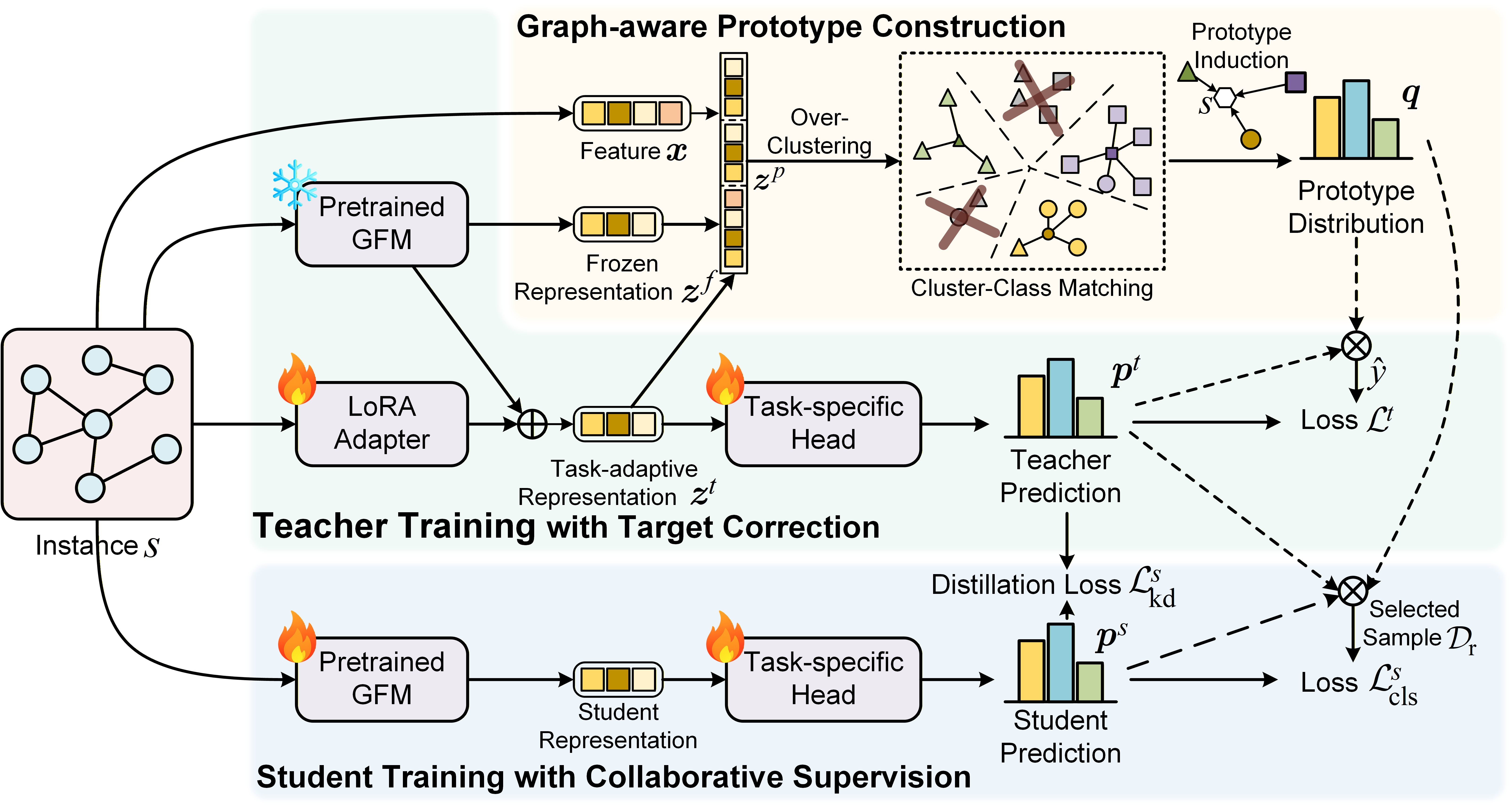}
\caption{Overview of PACT. During adaptation, PACT begins with graph-aware prototype construction, where a prototype-induced class distribution is derived from raw features, frozen GFM representations, and task-adaptive teacher representations. The resulting prototype evidence is then incorporated into teacher training with target correction for teacher refinement and reliable sample selection. Finally, in student training with collaborative supervision, the FFT student is optimized via soft distillation and reliability-aware hard supervision. Here, snowflake and flame icons indicate frozen and trainable modules, respectively; $\oplus$ denotes LoRA injection, and $\otimes$ denotes agreement-based operations.}
\label{fig:framework}
\end{figure*}

\section{Method}
In this section, we introduce our PACT framework for robust graph foundation model adaptation under label noise. As illustrated in Figure \ref{fig:framework}, PACT starts from a PEFT-FFT teacher-student architecture, which serves as the adaptation backbone to balance robustness and adaptation capacity (Section \ref{sec:architecture}). It then performs graph-aware prototype construction to induce class-level prototype evidence from raw features, frozen GFM representations, and task-adaptive teacher representations (Section \ref{sec:prototype}). Based on this prototype evidence, PACT further introduces teacher training with target correction and student training with collaborative supervision (Section \ref{sec:supervision}). Finally, we formulate the overall collaborative optimization objective and summarize the complete training procedure (Section \ref{sec:overall}).

\subsection{PEFT-FFT Teacher-Student Architecture} \label{sec:architecture}
Adapting graph foundation models under label noise presents a fundamental trade-off. Sufficient model updating is needed to align pre-trained representations with the downstream task, yet under noisy supervision, the same adaptation process can easily corrupt the transferable priors inherited from pre-training. 

To address this trade-off, PACT couples a PEFT teacher with an FFT student in a teacher-student framework. This design is motivated by the empirical complementarity between PEFT and FFT under noisy adaptation: PEFT better preserves pre-trained priors under restricted parameter updates, while FFT offers stronger adaptation capacity. Specifically, given a pre-trained GFM encoder $f_{\theta}$ with pre-trained parameters $\theta$, PACT first instantiates two encoder branches: a teacher encoder $f_{\theta^t}$ with parameters $\theta^t$ and a student encoder $f_{\theta^s}$ with parameters $\theta^s$, where both $\theta^t$ and $\theta^s$ are initialized from $\theta$. A task-specific classifier head is then attached to each branch, yielding a PEFT teacher and an FFT student.

\smallskip\noindent\textbf{PEFT teacher architecture.} For the teacher branch, PACT adopts LoRA as the PEFT strategy, as its low-rank adaptation form limits rapid memorization of noisy labels, making it better suited for stable teacher supervision under label noise \cite{steele2026lora}. Starting from the pre-trained backbone $f_{\theta^t}$, PACT freezes the backbone parameters $\theta^t$ and injects trainable low-rank updates into selected linear transformations. Concretely, for a linear layer with weight matrix $W \in \mathbb{R}^{d_{\mathrm{out}} \times d_{\mathrm{in}}}$, where $d_{\mathrm{in}}$ and $d_{\mathrm{out}}$ denote the input and output dimensions of the layer, LoRA parameterizes the adapted weight as
\begin{equation}
    W' = W + \frac{\alpha}{r}EF,
\end{equation}
where $W$ denotes the frozen pre-trained weight inherited from $\theta^t$, $F \in \mathbb{R}^{r \times d_{\mathrm{in}}}$ and $E \in \mathbb{R}^{d_{\mathrm{out}} \times r}$ are trainable low-rank matrices, $r \ll \min(d_{\mathrm{in}}, d_{\mathrm{out}})$ is the rank hyperparameter controlling the capacity of the low-rank update, and $\alpha$ is a scaling hyperparameter that controls the update magnitude.

Let $\theta^t_{\mathrm{lora}}$ denote all LoRA parameters and $g_{\phi^t}$ denote the teacher classification head. Then, built upon the frozen backbone parameters $\theta^t$, the teacher predictor is defined as
\begin{equation}
h^t(s)=g_{\phi^t}\big(f_{(\theta^t,\theta^t_{\mathrm{lora}})}(A_s,X_s)\big),
\end{equation}
where $(A_s,X_s)$ denotes the contextual graph view associated with instance $s$. Here, the backbone parameters $\theta^t$ are frozen, while the LoRA and head parameters $\{\theta^t_{\mathrm{lora}},\phi^t\}$ are trainable.

\smallskip\noindent\textbf{FFT student architecture.} The student branch performs full fine-tuning for maximum adaptation capacity. Starting from the pre-trained backbone $f_{\theta^s}$, the student branch attaches a classification head $g_{\phi^s}$ and defines its predictor as
\begin{equation}
    h^s(s)=g_{\phi^s}\big(f_{\theta^s}(A_s,X_s)\big),
\end{equation}
where all student parameters $\{\theta^s,\phi^s\}$ are updated during adaptation.

By combining these two branches, PACT establishes a collaborative architecture that couples stable teacher supervision with high-capacity student adaptation. In the following sections, we describe how this architecture is further integrated with prototype calibration for robust adaptation under label noise.

\subsection{Graph-aware Prototype Construction} \label{sec:prototype}
With the PEFT-FFT teacher-student architecture in place, PACT obtains a more stable adaptation basis in which pre-trained knowledge is less likely to be rapidly distorted by noisy supervision. Building on this basis, we further introduce a graph-aware prototype construction module to organize the preserved priors into class-level references. A straightforward solution would be to construct prototypes directly from the current adapted representations. However, such a strategy may still inherit the bias introduced by corrupted supervision. To reduce this risk, PACT constructs prototypes by anchoring them to adaptation-insulated frozen representations that preserve the structural priors captured during pre-training, while also incorporating task-relevant cues from the PEFT teacher under graph structure constraints.

\smallskip\noindent\textbf{Prototype representation construction.} 
We begin by constructing a dedicated representation for prototype induction that combines the preserved geometric priors of the frozen GFM backbone with the task-adaptive cues provided by the PEFT teacher. Given an instance $s$ with its contextual graph view $(A_s,X_s)$, we characterize it from three complementary sources for reliable prototype induction: 
(1) its raw attribute feature $\boldsymbol{x}$, which remains uninfluenced by any supervision;
(2) its frozen geometric representation $\boldsymbol{z}^{f}=f_{\theta}(A_s,X_s)$, which preserves source-domain structural priors while remaining insulated from noisy labels; and
(3) its task-adaptive representation $\boldsymbol{z}^{t}=f_{(\theta^t,\theta^t_{\mathrm{lora}})}(A_s,X_s)$, which reflects target-task information through the robust PEFT teacher. With $\rho(\cdot)$ denoting $\ell_2$ normalization, the composite prototype representation for instance $s$ is constructed as
\begin{equation}
\boldsymbol{z}^{p}
=
\rho \!\left(
\left[
\rho(\boldsymbol{x}),\,
\rho(\boldsymbol{z}^{f}),\,
\rho(\boldsymbol{z}^{t})
\right]
\right),
\end{equation}
where $[\cdot,\cdot,\cdot]$ denotes feature concatenation. By doing so, PACT represents each instance in a way that is simultaneously anchored to supervision-free attributes, noise-insulated pre-trained geometry, and robust task-adaptive structure, thereby providing a more reliable basis for subsequent prototype induction.

\smallskip\noindent\textbf{Prototype induction.}
After obtaining the robust prototype representation $\boldsymbol{z}^{p}$ for each training instance, PACT constructs class prototypes in a label-decoupled manner. Instead of directly computing class centroids from observed label groups, which can be corrupted by mislabeled instances, we first perform label-agnostic over-clustering on prototype representation space, since instances from the same semantic class may still exhibit multiple local modes under noisy adaptation. We then recover semantic correspondence through cluster-class matching to obtain class-level prototypes.

Specifically, let $\mathcal{S}_{\mathrm{tr}}$ denote the set of training instances used for prototype construction, and let $M>C$ be the number of over-clusters. We first partition $\{\boldsymbol{z}^{p}\mid s\in\mathcal{S}_{\mathrm{tr}}\}$ into $M$ clusters via $k$-means, yielding a cluster assignment map $\pi(s)\in\{1,\dots,M\}$. To associate these clusters with semantic classes, we then inject teacher-side robust soft class information into each cluster. Let $\boldsymbol{p}^{t}_s=\mathrm{softmax}\big(h^{t}(s)\big)\in\mathbb{R}^{C}$
denote the teacher predictive distribution for instance $s$. For each cluster $\mathcal{C}_m=\{\,s\in\mathcal{S}_{\mathrm{tr}} \mid \pi(s)=m\,\}$, we compute its class score by averaging teacher predictions over the assigned instances:
\begin{equation}
\boldsymbol{u}_m=\frac{1}{|\mathcal{C}_m|}\sum_{s\in\mathcal{C}_m}\boldsymbol{p}^{t}_s.
\end{equation}
Collecting all cluster-level scores yields a matrix $U\in\mathbb{R}^{M\times C}$, whose $m$-th row is given by $\boldsymbol{u}_m^\top$.

To transform these fine-grained clusters into class-level prototypes, we introduce a binary assignment matrix $B\in\{0,1\}^{M\times C}$, where $B_{mc}=1$ indicates that cluster $\mathcal{C}_m$ is matched to class $c$. The matching is determined by
\begin{equation}
\begin{aligned}
&\max_{B}\ \sum_{m=1}^{M}\sum_{c=1}^{C} B_{mc}U_{mc}\\
\text{s.t.}
&\sum_{c'=1}^{C}B_{mc'}\leq 1,\ \forall m\\
&\sum_{m'=1}^{M}B_{m'c}\leq 1,\ \forall c.
\end{aligned}
\end{equation}
where $U_{mc}$ denotes the $(m,c)$-th entry of $U$, representing the semantic score of assigning cluster $\mathcal{C}_m$ to class $c$. In implementation, we approximately solve this assignment problem using greedy matching. Specifically, at each step, the cluster-class pair with the highest semantic score is selected, and the corresponding cluster and class are excluded from subsequent matching until all classes are matched or no valid pair remains.

The above constraints enforce a one-to-one matching between classes and representative clusters, while the objective favors assignments with high overall semantic consistency. As a result, over-clustering allows PACT to preserve the most semantically representative fine-grained structures for prototype extraction, while leaving unmatched clusters outside subsequent class-level aggregation. For each class $c$, we then compute its prototype by averaging the prototype representations of the instances within its matched cluster, i.e., $\boldsymbol{p}_c=\frac{1}{|\mathcal{C}_c|}\sum_{s\in\mathcal{C}_c}\boldsymbol{z}_s^{p},$ where $\mathcal{C}_c$ denotes the representative cluster assigned to class $c$, and $\boldsymbol{z}_s^{p}$ denotes prototype representation $\boldsymbol{z}^{p}$ for instance $s$. By doing so, on the one hand, the one-to-one constraint assigns different prototype anchors to different classes, preventing multiple classes from collapsing onto the same dominant cluster. On the other hand, since over-clustering produces more clusters than classes, only clusters with stronger class correspondence are selected as prototype anchors, while unmatched clusters, including potentially noise-induced abnormal clusters, are excluded from the final class prototype construction. Together, these properties yield more reliable class-level references for subsequent calibration.

\subsection{Prototype-calibrated Collaborative Supervision} \label{sec:supervision}
Building upon the constructed graph-aware class prototypes, PACT further uses them to recalibrate the noisy supervision used during adaptation. The key idea is to first convert the prototype space into instance-wise soft class evidence, and then combine it with the teacher prediction to form a corrected supervisory target. This corrected target is subsequently used to refine teacher optimization and to guide student learning.

\smallskip\noindent\textbf{Prototype-induced target correction.} 
To leverage the GFM-based prototypes for supervision calibration, we first derive instance-wise soft class evidence from the prototype-guided representation and then combine it with the teacher prediction to form a calibrated target. 

Specifically, 
given a training pair $(s,\tilde{y})\in\widetilde{\mathcal{D}}_{\mathrm{tr}}$, where $\widetilde{\mathcal{D}}_{\mathrm{tr}} \subset \widetilde{\mathcal{D}}$ denotes the noisy training set, we first transform its prototype-guided representation $\boldsymbol{z}^{p}$ into instance-wise class evidence by computing its similarity to each class prototype, yielding a prototype-induced class distribution $\boldsymbol{q}$. For each class $c\in\{1,2,\dots,C\}$, we define
\begin{equation}
    \boldsymbol{q}_{c}=
    \frac{
    \exp\!\left(\mathrm{sim}(\boldsymbol{z}^{p},\boldsymbol{p}_c)/\tau\right)
    }{
    \sum_{c'=1}^{C}
    \exp\!\left(\mathrm{sim}(\boldsymbol{z}^{p},\boldsymbol{p}_{c'})/\tau\right)
    },
\end{equation}
where $\boldsymbol{q}_c$ denotes the $c$-th component of $\boldsymbol{q}$, representing the prototype-induced probability of assigning $s$ to class $c$, $\mathrm{sim}(\cdot,\cdot)$ denotes cosine similarity, and $\tau$ is a temperature parameter.

Subsequently, we calibrate noisy supervision by leveraging the agreement between the prototype-induced class distribution and the teacher prediction. Let $\boldsymbol{p}^t=\mathrm{softmax}(h^t(s))$ denote the teacher predictive distribution for $s$. We define an agreement distribution over classes as
\begin{equation}
    \boldsymbol{a}
    =
    \frac{\boldsymbol{p}^{t}\odot \boldsymbol{q}}
    {\|\boldsymbol{p}^{t}\odot \boldsymbol{q}\|_1},
\end{equation}
where $\odot$ denotes the Hadamard product and $\|\cdot\|_1$ is the $\ell_1$ norm. Based on this agreement, the calibrated target is constructed as
\begin{equation}
\hat{\boldsymbol{y}}=
\begin{cases}
\boldsymbol{e}_{\tilde{y}}, & \text{if consensus holds},\\
\beta\,\boldsymbol{e}_{\tilde{y}}+(1-\beta)\,\boldsymbol{a}, & \text{otherwise},
\end{cases}
\end{equation}
where consensus holds when the observed label $\tilde{y}$ is the most probable class under both the teacher predictive distribution and the prototype-induced distribution, i.e., $\tilde{y}=\arg\max_c \boldsymbol{p}^t_c$ and $\tilde{y}=\arg\max_c \boldsymbol{q}_c$. Here, $\boldsymbol{e}_{\tilde{y}}$ denotes the one-hot vector of the observed label $\tilde{y}$, and $\beta$ is an interpolation weight controlling the trade-off between the original noisy label and the agreement-induced soft target. In practice, we set $\beta=\sqrt{\boldsymbol{p}^t_{\tilde{y}}\,\boldsymbol{q}_{\tilde{y}}}$ to reflect their joint confidence in the observed label $\tilde{y}$. In this way, the calibrated target preserves reliable noisy supervision when consensus is strong, while softly correcting potentially corrupted labels according to the shared belief of the teacher and the prototypes.

Then, the teacher model $h^t$ is optimized using the calibrated target via the following soft cross-entropy classification loss:
\begin{equation}
\mathcal{L}^{t}
=
-\frac{1}{|\widetilde{\mathcal{D}}_{\mathrm{tr}}|}
\sum_{(s,\tilde{y})\in\widetilde{\mathcal{D}}_{\mathrm{tr}}}
\sum_{c=1}^{C}\hat{y}_{c}\log \Big(\mathrm{softmax}(h^{t}(s))_{c}\Big),
\label{eq:lteacher}
\end{equation}
where $\mathrm{softmax}(h^{t}(s))_{c}$ denotes the teacher predicted probability for instance $s$ of class $c$. This allows the teacher to absorb prototype-aware label corrections during adaptation.

\subsection{Overall Collaborative Optimization} \label{sec:overall}
\begin{algorithm}[t]
\caption{Training Procedure of PACT}
\label{alg:pact}
\begin{algorithmic}[1]
\REQUIRE Noisy training set $\widetilde{\mathcal{D}}_{\mathrm{tr}}$, pre-trained GFM encoder, epochs $T$, temperatures $\tau,\tau_2$, balance weight $\lambda$
\ENSURE Adapted student model $h^s$ (and teacher model $h^t$)

\STATE Initialize a PEFT teacher $h^t$ and an FFT student $h^s$ from the same pre-trained GFM
\FOR{$e=1$ \TO $T$}
    \STATE Construct graph-aware class prototypes $\{\boldsymbol{p}_c\}_{c=1}^{C}$
    \FORALL{$(s,\tilde{y})\in\widetilde{\mathcal{D}}_{\mathrm{tr}}$}
        \STATE Compute the prototype-induced distribution $\boldsymbol{q}$
        \STATE Compute the teacher prediction $\boldsymbol{p}^{t}$
        \STATE Construct the calibrated target $\hat{\boldsymbol{y}}$
    \ENDFOR
    \STATE Update the teacher by minimizing loss $\mathcal{L}^{t}$ according to Eq. \eqref{eq:lteacher}
    \STATE Compute the teacher-filtered distillation loss $\mathcal{L}_{\mathrm{kd}}^{s}$ according to Eq. \eqref{eq:skd}
    \STATE Construct the refined subset $\mathcal{D}_{\mathrm{r}}$ according to Eq. \eqref{eq:Dr}
    \STATE Compute the reliability-aware supervision loss $\mathcal{L}_{\mathrm{cls}}^{s}$ according to Eq. \eqref{eq:scls}
    \STATE Update the student by minimizing loss $\mathcal{L}^{s}=\mathcal{L}_{\mathrm{kd}}^{s}+\lambda \mathcal{L}_{\mathrm{cls}}^{s}$ according to Eq. \eqref{eq:lstudent}
\ENDFOR
\RETURN $h^s$ and $h^t$
\end{algorithmic}
\end{algorithm}

\smallskip\noindent\textbf{Teacher-filtered distillation with reliable supervision.}
Although the calibrated target is suitable for refining the teacher, directly using it to supervise the student is less desirable, as the student is more prone to overfitting residual noise and its evolving GFM backbone may cause representation drift. We therefore adopt teacher-filtered soft distillation as the main guidance for student adaptation, leveraging the teacher as an additional noise-filtering layer after prototype-calibrated refinement. In addition, we incorporate reliability-aware hard supervision on a refined sample set to provide trustworthy label anchors.

Let $\boldsymbol{p}^{s}=\mathrm{softmax}(h^{s}(s))$ denote the student predictive distribution. For teacher-filtered soft distillation, we distill from the teacher prediction, where the teacher itself has been refined by the above prototype-aware target correction, so that the student is guided by a more stable and noise-filtered soft signal. The corresponding distillation loss is defined as 
\begin{equation}
\mathcal{L}_{\mathrm{kd}}^{s}
=
\frac{\tau_2^{2}}{|\widetilde{\mathcal{D}}_{\mathrm{tr}}|}
\sum_{(s,\tilde{y})\in\widetilde{\mathcal{D}}_{\mathrm{tr}}}
\mathrm{KL}
\!\left(
\sigma_{\tau_2}(h^{t}(s))
\;\middle\|\;
\sigma_{\tau_2}(h^{s}(s))
\right),
\label{eq:skd}
\end{equation}
where $\sigma_{\tau_2}(h(s))_{c}
=
\frac{\exp(h_{c}(s)/\tau_2)}
{\sum_{c'=1}^{C}\exp(h_{c'}(s)/\tau_2)}$
denotes the softmax output with distillation temperature $\tau_2$.

To complement the soft guidance provided by teacher-filtered distillation, we further perform agreement-aware sample selection to identify a refined subset, on which hard supervision is then imposed. Formally, we construct 
\begin{equation}
\begin{aligned}
\mathcal{D}_{\mathrm{r}}=\{
(s,\tilde{y}) \in \widetilde{\mathcal{D}}_{\mathrm{tr}} \mid &\arg\max_c \boldsymbol{p}^{t}_{c}=\arg\max_c \boldsymbol{p}^{s}_{c}=\\
& \arg\max_c \boldsymbol{q}_{c}=\tilde{y},\\
&\max_c \boldsymbol{p}^{t}_{c} \ge \eta,\quad
\max_c \boldsymbol{p}^{s}_{c} \ge \eta
\},
\label{eq:Dr}
\end{aligned}
\end{equation}
where $\eta$ is a confidence threshold that retains only reliable samples for hard supervision of the student. The corresponding classification loss on selected samples is defined as 
\begin{equation}
\mathcal{L}_{\mathrm{cls}}^{s}
=
-\frac{1}{|\mathcal{D}_{\mathrm{r}}|}
\sum_{(s,\tilde{y})\in\mathcal{D}_{\mathrm{r}}}
\log \Big(\mathrm{softmax}(h^{s}(s))_{\tilde{y}}\Big),
\label{eq:scls}
\end{equation}
where $\mathrm{softmax}(h^{s}(s))_{\tilde{y}}$ denotes the student predicted probability for instance $s$ of class $\tilde{y}$. This provides explicit supervision on samples with four-way agreement among the observed label, the teacher prediction, the student prediction, and the prototype-induced prediction, thereby offering more reliable label anchors to complement the teacher-filtered soft guidance.

Combining the above components, PACT performs collaborative optimization of the teacher and student under prototype-calibrated supervision. As summarized in Algorithm~\ref{alg:pact}, within each training epoch, PACT first constructs graph-aware class prototypes (line 3), and then derives prototype-induced class evidence and calibrated targets for noisy training pairs (lines 4--8). Based on the agreement between the prototype-induced distribution and the teacher prediction, the teacher is refined via $\mathcal{L}^{t}$ (line 9), allowing it to absorb prototype-aware label corrections during adaptation.

After teacher refinement, the student is optimized under two complementary supervisory signals. The primary supervision is teacher-filtered soft distillation over the full noisy training set, through which the prototype-calibrated teacher provides a more stable and noise-filtered soft signal (line 10). In addition, reliability-aware hard supervision is imposed on the refined subset $\mathcal{D}_{\mathrm{r}}$ (lines 11--12), whose labels are jointly supported by the observed label, the teacher prediction, the student prediction, and the prototype-induced prediction. The overall student objective is defined as
\begin{equation}
\mathcal{L}^{s}
=
\mathcal{L}_{\mathrm{kd}}^{s}
+
\lambda\,\mathcal{L}_{\mathrm{cls}}^{s},
\label{eq:lstudent}
\end{equation}
where $\lambda$ balances the contribution of the reliable hard supervision term (line 13). In this way, the PEFT teacher serves as a prototype-calibrated noise filter, while the FFT student benefits from both stable soft guidance and explicit hard label anchors on four-way consistent samples.

During inference, PACT uses only the fully fine-tuned student for prediction. The PEFT teacher and prototype-related components are used only during training and introduce no additional inference-time overhead.

\begin{table}[t]
\centering
\caption{Statistics of the datasets used in the experiments.}
\label{tab:dataset_stats}
\small
\setlength{\tabcolsep}{5pt}
\resizebox{\linewidth}{!}{
\begin{tabular}{lrrrrrrr}
\toprule
Dataset & \#Nodes & \#Edges & \#Feat. & \#Cls. & \#Train & \#Val & \#Test \\
\midrule
Cora & 2,708 & 10,556 & 1,433 & 7 & 140 & 500 & 1,000 \\
Citeseer & 3,327 & 9,104 & 3,703 & 6 & 120 & 500 & 1,000 \\
PubMed & 19,717 & 88,648 & 500 & 3 & 60 & 500 & 1,000 \\
Airports & 1,190 & 13,599 & 1,190 & 4 & 833 & 119 & 238 \\
Clothing1M-graph & 15,524 & 146,666 & 1,000 & 14 & 4,498 & 500 & 10,526 \\
\bottomrule
\end{tabular}
}
\end{table}

\begin{table*}[t]
\centering
\caption{Performance comparison of different adaptation strategies under various label noise settings. The best results are in \textbf{bold}, and the second-best results are \underline{underlined}.}
\label{tab:riemanngfm_nc_adaptation}
\begin{tabular}{llccccccc}
\toprule
\multirow{2}{*}{Dataset} & \multirow{2}{*}{Method} & \multirow{2}{*}{Clean} 
& \multicolumn{2}{c}{Uniform Noise} 
& \multicolumn{2}{c}{Pair Noise} 
& \multicolumn{2}{c}{IDN Noise} \\
\cmidrule(lr){4-5} \cmidrule(lr){6-7} \cmidrule(lr){8-9}
& & & 20\% & 50\% & 20\% & 40\% & 20\% & 50\% \\
\midrule

\multirow{7}{*}{Cora}
& LoRA & 80.52 ± 0.72 & \underline{68.62 ± 2.97} & \underline{52.62 ± 2.16} & 67.32 ± 1.68 & 45.62 ± 3.42 & 71.04 ± 1.71 & 31.72 ± 12.08 \\
& FFT & \textbf{81.00 ± 2.01} & 63.88 ± 1.31 & 48.78 ± 0.70 & 67.08 ± 2.93 & \underline{47.86 ± 1.84} & 72.94 ± 2.60 & 33.59 ± 16.54 \\
& NRGNN & 77.40 ± 1.85 & 60.32 ± 6.17 & 31.66 ± 9.65 & \underline{71.38 ± 3.79} & 41.14 ± 4.77 & 69.24 ± 3.87 & 53.96 ± 8.83 \\
& ProCon & 66.38 ± 1.46 & 64.82 ± 5.51 & 43.24 ± 11.03 & 67.82 ± 3.13 & 47.18 ± 3.05 & 72.90 ± 2.17 & 30.18 ± 9.39 \\
& Forward & \underline{80.92 ± 0.57} & 64.98 ± 2.36 & 34.14 ± 10.11 & 70.00 ± 2.70 & 44.80 ± 2.97 & \underline{73.94 ± 1.69} & 33.16 ± 10.72 \\
& Co-teaching+ & 75.57 ± 1.75 & 62.69 ± 4.71 & 46.84 ± 5.22 & 64.86 ± 1.42 & 40.41 ± 3.97 & 65.08 ± 4.02 & \underline{54.65 ± 4.54} \\
& PACT & 80.82 ± 0.39 & \textbf{70.78 ± 3.34} & \textbf{55.28 ± 4.50} & \textbf{71.94 ± 1.78} & \textbf{47.90 ± 2.98} & \textbf{75.82 ± 1.15} & \textbf{58.64 ± 2.03} \\
\midrule

\multirow{7}{*}{Citeseer}
& LoRA & 66.22 ± 1.35  & 58.00 ± 2.57 & 30.92 ± 2.97 & 54.10 ± 1.83 & \underline{35.26 ± 3.47} & 57.42 ± 1.20 & 45.10 ± 3.58 \\
& FFT &  66.66 ± 0.35  & 58.24 ± 2.33 & 28.66 ± 5.11 & 52.90 ± 1.88 & 32.72 ± 1.63 & 60.85 ± 1.75 & 36.92 ± 4.12 \\
& NRGNN & \underline{67.18 ± 1.00} & \underline{59.36 ± 1.29} & 32.22 ± 7.72 & \underline{54.94 ± 1.00} & 33.02 ± 3.19 & 59.32 ± 3.48 & 40.78 ± 5.19 \\
& ProCon & 66.38 ± 1.46 & 56.86 ± 3.04 & 38.02 ± 6.41 & 53.92 ± 1.07 & 34.14 ± 1.22 & \underline{62.20 ± 1.81} & 39.34 ± 2.69 \\
& Forward & 65.86 ± 1.58 & 58.38 ± 2.29 & 30.24 ± 8.61 & 52.92 ± 1.47 & 33.78 ± 1.02 & 61.36 ± 3.04 & 40.38 ± 4.71 \\
& Co-teaching+ & 64.05 ± 1.81 & 52.84 ± 2.31 & \underline{39.26 ± 2.97} & 50.87 ± 1.46 & 33.44 ± 1.83 & 53.54 ± 2.18 & \underline{45.65 ± 4.73} \\
& PACT & \textbf{68.22 ± 2.15} & \textbf{60.94 ± 2.36} & \textbf{48.24 ± 2.10} & \textbf{57.40 ± 0.95} & \textbf{37.62 ± 2.12} & \textbf{64.14 ± 2.55} & \textbf{53.80 ± 1.65} \\
\midrule

\multirow{7}{*}{PubMed}
& LoRA  & 73.36 ± 6.75 & 61.32 ± 3.49 & 48.76 ± 3.37 & \underline{64.35 ± 11.92} & \underline{55.72 ± 9.28} & 63.72 ± 4.23 & 26.58 ± 8.14 \\
& FFT  & \textbf{77.52 ± 0.76} & \underline{63.94 ± 5.16} & 39.98 ± 4.36 & 57.46 ± 6.99 & 55.22 ± 6.58 & 65.52 ± 4.06 & 30.18 ± 10.88 \\
& NRGNN  & 47.80 ± 11.55 & 54.24 ± 10.34 & \textbf{53.20 ± 6.14} & 45.44 ± 9.67 & 51.68 ± 7.75 & 55.84 ± 8.78 & 24.60 ± 10.79 \\
& ProCon & \underline{77.52 ± 0.91} & 61.40 ± 3.54 & 43.60 ± 5.42 & 48.98 ± 9.81 & 50.20 ± 7.64 & \underline{65.82 ± 4.07} & \underline{30.28 ± 13.01} \\
& Forward & 77.40 ± 1.21 & 57.46 ± 8.22 & 46.28 ± 4.20 & 48.98 ± 10.59 & 51.92 ± 3.65 & 63.10 ± 5.10 & 28.16 ± 10.66 \\
& Co-teaching+ & 66.78 ± 4.91 & 52.02 ± 7.10 & 43.73 ± 3.93 & 50.89 ± 9.49 & 37.42 ± 5.06 & 46.21 ± 5.41 & 27.77 ± 8.56 \\
& PACT  & 76.72 ± 1.36 & \textbf{66.60 ± 5.06} & \underline{50.66 ± 2.15} & \textbf{70.14 ± 1.54} & \textbf{59.98 ± 3.78} & \textbf{69.06 ± 2.55} & \textbf{46.82 ± 1.85} \\
\midrule

\multirow{7}{*}{Airports}
& LoRA & 54.03 ± 3.47 & \underline{53.78 ± 1.25} & 39.16 ± 6.26 & 52.10 ± 0.84 & 32.44 ± 4.90 & 54.03 ± 2.77 & 40.67 ± 10.09 \\
& FFT &  55.80 ± 2.62  & 48.15 ± 4.37 & 37.23 ± 3.73 & 51.01 ± 2.56 & \underline{39.92 ± 7.33} & 54.12 ± 1.31 & 32.35 ± 9.80 \\
& NRGNN & 53.28 ± 3.48  & 47.31 ± 2.26 & 40.08 ± 9.69 & 45.21 ± 3.61 & 26.55 ± 6.23 & 48.32 ± 2.38 & 37.48 ± 6.30 \\
& ProCon & \underline{57.48 ± 1.31} & 50.67 ± 2.77 & \underline{46.64 ± 2.99} & \textbf{55.21 ± 2.03} & 36.39 ± 6.45 & \underline{56.64 ± 3.60} & \underline{46.13 ± 2.10} \\
& Forward & 56.97 ± 1.11 & 52.86 ± 4.25 & 41.26 ± 7.60 & \underline{53.53 ± 2.09} & 36.81 ± 7.33 & 55.29 ± 2.82 & 44.12 ± 3.49 \\
& Co-teaching+ & 53.78 ± 4.66 & 37.98 ± 12.81 & 36.39 ± 4.48 & 48.66 ± 2.44 & 37.52 ± 9.07 & 52.90 ± 1.26 & 34.75 ± 4.74 \\
& PACT & \textbf{57.56 ± 1.03}  & \textbf{56.39 ± 2.48} & \textbf{49.16 ± 4.73} & 53.03 ± 2.31 & \textbf{45.97 ± 5.00} & \textbf{57.31 ± 1.15} & \textbf{46.47 ± 1.61} \\
\bottomrule
\end{tabular}
\end{table*}

\begin{table}[t]
\centering
\caption{Results on Clothing1M-graph.}
\label{tab:clothing1m_results}
\resizebox{\linewidth}{!}{
\begin{tabular}{ccccccc}
\toprule
LoRA & FFT & Forward & Co-teaching+ & NRGNN & ProCon & PACT \\
\midrule
11.27$\pm$1.05 & 11.02$\pm$1.05 & 11.06 ± 2.61 & 8.53±1.22 & 7.60$\pm$2.24 & \underline{11.67$\pm$0.68} & \textbf{13.49$\pm$8.03} \\
\bottomrule
\end{tabular}
}
\end{table}

\begin{table}[t]
\centering
\caption{Few-shot node classification results under lower label budgets. The best results are in \textbf{bold}.}
\label{tab:few_shot_results}
\setlength{\tabcolsep}{4pt}
\begin{tabular}{llccc}
\toprule
\multicolumn{5}{c}{5-shot}\\
\midrule
Dataset & Method & Uniform-50\% & Pair-40\% & IDN-50\% \\
\midrule
\multirow{3}{*}{Cora}
& NRGNN  & 23.14 ± 4.93 & \textbf{50.04 ± 7.22} & 22.76 ± 3.46 \\
& ProCon & 20.50 ± 2.89 & 49.08 ± 4.72 & 24.22 ± 7.04 \\
& PACT   & \textbf{34.72 ± 3.90} & 45.60 ± 1.94 & \textbf{29.34 ± 9.06} \\
\midrule
\multirow{3}{*}{Citeseer}
& NRGNN  & 26.52 ± 7.45 & 34.84 ± 4.02 & 26.26 ± 5.71 \\
& ProCon & 24.56 ± 3.53 & 37.38 ± 2.84 & 14.00 ± 6.63 \\
& PACT   & \textbf{30.60 ± 2.46} & \textbf{42.52 ± 1.63} & \textbf{34.08 ± 2.64} \\
\midrule
\multirow{3}{*}{PubMed}
& NRGNN  & 43.84 ± 3.78 & 34.26 ± 10.20 & 28.30 ± 7.19 \\
& ProCon & 38.00 ± 5.16 & 35.12 ± 5.90 & 26.22 ± 8.19 \\
& PACT   & \textbf{50.44 ± 4.79} & \textbf{40.76 ± 3.83} & \textbf{46.58 ± 1.97} \\
\midrule
\multirow{3}{*}{Airports}
& NRGNN  & 23.94 ± 4.92 & 25.84 ± 0.46 & 25.58 ± 7.01 \\
& ProCon & 24.63 ± 7.11 & 26.58 ± 3.27 & \textbf{25.84 ± 3.60} \\
& PACT   & \textbf{24.93 ± 2.02} & \textbf{26.98 ± 1.96} & 24.91 ± 6.16 \\
\midrule
\midrule

\multicolumn{5}{c}{10-shot}\\
\midrule
Dataset & Method & Uniform-50\% & Pair-40\% & IDN-50\% \\
\midrule
\multirow{3}{*}{Cora}
& NRGNN  & 29.12 ± 5.53 & 34.66 ± 1.45 & 21.58 ± 7.43 \\
& ProCon & 19.20 ± 7.96 & 43.32 ± 3.62 & 20.82 ± 7.06 \\
& PACT   & \textbf{40.86 ± 4.19} & \textbf{43.38 ± 3.18} & \textbf{39.64 ± 1.18} \\
\midrule
\multirow{3}{*}{Citeseer}
& NRGNN  & 32.16 ± 8.04 & 36.18 ± 4.09 & 21.22 ± 7.40 \\
& ProCon & 23.86 ± 3.93 & 35.72 ± 1.77 & 12.80 ± 7.14 \\
& PACT   & \textbf{35.44 ± 0.77} & \textbf{36.90 ± 1.08} & \textbf{33.22 ± 1.83} \\
\midrule
\multirow{3}{*}{PubMed}
& NRGNN  & 38.80 ± 5.47 & 44.46 ± 6.33 & 38.04 ± 2.62 \\
& ProCon & 38.44 ± 3.31 & 48.42 ± 5.81 & \textbf{42.70 ± 5.16} \\
& PACT   & \textbf{45.40 ± 7.61} & \textbf{53.38 ± 8.80} & 42.28 ± 6.35 \\
\midrule
\multirow{3}{*}{Airports}
& NRGNN  & \textbf{29.23 ± 3.11} & 25.66 ± 2.18 & \textbf{26.48 ± 1.30} \\
& ProCon & 28.48 ± 2.60 & 29.10 ± 2.00 & 25.94 ± 0.93 \\
& PACT   & 22.70 ± 3.54 & \textbf{29.35 ± 2.32} & 24.33 ± 2.69 \\
\midrule
\midrule

\multicolumn{5}{c}{15-shot}\\
\midrule
Dataset & Method & Uniform-50\% & Pair-40\% & IDN-50\% \\
\midrule
\multirow{3}{*}{Cora}
& NRGNN  & 35.62 ± 12.98 & 46.56 ± 6.33 & 30.14 ± 10.92 \\
& ProCon & 45.52 ± 2.96 & 47.86 ± 4.53 & 29.60 ± 8.30 \\
& PACT   & \textbf{50.22 ± 4.44} & \textbf{47.94 ± 2.38} & \textbf{47.98 ± 2.77} \\
\midrule
\multirow{3}{*}{Citeseer}
& NRGNN  & 22.72 ± 2.04 & 33.44 ± 3.81 & 28.80 ± 5.46 \\
& ProCon & 25.80 ± 2.72 & 33.10 ± 1.52 & 27.90 ± 3.87 \\
& PACT   & \textbf{38.42 ± 3.06} & \textbf{36.36 ± 2.54} & \textbf{42.68 ± 2.77} \\
\midrule
\multirow{3}{*}{PubMed}
& NRGNN  & 39.40 ± 9.49 & 48.56 ± 7.23 & 37.74 ± 10.34 \\
& ProCon & 26.64 ± 5.71 & 57.30 ± 2.77 & 29.80 ± 8.39 \\
& PACT   & \textbf{40.86 ± 4.04} & \textbf{65.14 ± 0.53} & \textbf{47.76 ± 4.70} \\
\midrule
\multirow{3}{*}{Airports}
& NRGNN  & 29.41 ± 4.80 & 38.64 ± 1.67 & 32.03 ± 4.10 \\
& ProCon & 26.49 ± 1.87 & 38.20 ± 2.80 & 33.81 ± 4.55 \\
& PACT   & \textbf{29.48 ± 5.26} & \textbf{38.99 ± 1.14} & \textbf{39.07 ± 1.86} \\
\bottomrule
\end{tabular}

\end{table}

\section{Experiments}
\subsection{Experimental Setup}
\smallskip\noindent\textbf{Datasets.} We evaluate the adaptation performance of PACT on four widely used graph benchmark datasets (Cora, Citeseer, PubMed \cite{sen2008collective}, and Airports \cite{ribeiro2017struc2vec}) and one real-world noisy dataset (Clothing1M \cite{xiao2015learning}). Following common practice \cite{dai2021nrgnn, qian2023robust}, we manually inject three types of label noise into graph benchmark datasets. For Clothing1M, we sample 4,998 instances for training/validation following \cite{xia2023gnn,zhao2024training} and use the original test set as test nodes, constructing a 5-nearest-neighbor graph for graph-based learning with node features extracted by an ImageNet-pretrained ResNet-50 \cite{he2016deep} (denoted as Clothing1M-graph). We mainly use RiemannGFM \cite{sun2025riemanngfm} as the backbone to evaluate the effectiveness of the proposed adaptation method, and further assess its generalization across different backbones using SAMGPT \cite{yu2025samgpt}. Specifically, RiemannGFM is pre-trained on OGBN-Arxiv \cite{hu2020open}, Computers, and Physics \cite{shchur2018pitfalls}, while SAMGPT is pre-trained on Citeseer, PubMed, Photo \cite{shchur2018pitfalls}, and Computers. For downstream node classification, we follow the dataset split protocol used in \cite{sun2025riemanngfm}. For link prediction, we split Cora and Citeseer into training, validation, and test sets with a ratio of 5\%-5\%-90\%, while for PubMed and Airports, we use a split ratio of 0.45\%-0.05\%-99.5\%. Considering that clean labels are usually unavailable in real-world noisy-label scenarios, we follow the common noisy-label evaluation setting \cite{zhao2024estimating} and apply the same noisy-label setting to both the training and validation sets for all methods. Model selection and hyperparameter tuning are performed based on validation macro-F1 computed with the noisy validation labels, while clean labels are used only in the test set for final evaluation. The statistics for these datasets are shown in Table \ref{tab:dataset_stats}.

\smallskip\noindent\textbf{Synthetic noise injection.} We consider three synthetic label noise patterns that mimic different types of annotation errors: \textit{uniform flip noise}, \textit{pair flip noise}, and \textit{instance-dependent noise} (IDN). Uniform flip noise models fully random annotation mistakes, where a label may be flipped uniformly to any other class with specific noise rate \cite{patrini2017making}. Pair flip noise captures structured mistakes between confusing or semantically similar classes, where each class can only be corrupted into a predefined paired class under the specific noise rate \cite{han2018co}. IDN reflects instance-specific annotation uncertainty caused by factors such as ambiguous features or hard-to-distinguish samples; accordingly, each instance is assigned an individual flip probability determined by its features, and the corrupted label is then sampled based on that probability, following \cite{xia2020part}.

\smallskip\noindent\textbf{Baselines.}
We compare PACT with six representative baselines from three related perspectives, including standard GFM adaptation strategies, general noisy label learning methods, and graph-specific noisy-label learning approaches: (1) standard adaptation baselines: \textit{FFT} fully fine-tunes all parameters of the pre-trained GFM together with the task head (dual-layer GCN \cite{kipf2016semi}); \textit{LoRA} \cite{hu2022lora} adopts parameter-efficient fine-tuning by injecting trainable low-rank adapters into the pre-trained backbone while keeping most backbone parameters frozen; (2) general noisy label learning baselines: \textit{Forward} is a loss-correction method that alleviates label noise by modeling the label transition process during training; \textit{Co-teaching+} trains two peer networks simultaneously and lets them exchange selected samples during optimization, so as to reduce the impact of noisy labels through disagreement-aware collaborative training; (3) graph noisy-label learning baselines: \textit{NRGNN} is a representative graph noisy-label learning method that improves robustness by exploiting graph structure to identify and mitigate corrupted supervision; \textit{ProCon} is a recent strong baseline for graph learning with noisy labels, which enhances noise robustness through prototype- and contrastive-based regularization. For fair comparison, all baselines are implemented as adaptation strategies on the same pre-trained GFM for downstream tasks, with the same classifier architecture.

\smallskip\noindent\textbf{Implementation details.} For a fair comparison, we use a two-layer GCN classifier as the adaptation head for all methods, consistent with RiemannGFM. To construct the PEFT teacher, we inject LoRA adapters into selected linear layers of the GFM backbone. Specifically, for RiemannGFM, we set $r=8$ and $\alpha=16$ for the projection layers and the Euclidean initialization encoder, $r=2$ and $\alpha=4$ for the hyperbolic and spherical initialization encoders, and $r=4$ and $\alpha=8$ for the hyperbolic and spherical structure learners, with a dropout rate of $0.05$ for all LoRA adapters. The original parameters are frozen, and only the LoRA and the classification head parameters are optimized during teacher training. All models are optimized with Adam. Hyperparameters are selected based on validation macro-F1 from predefined candidate sets: the learning rate from $\{1 \times 10^{-1}, 5 \times 10^{-2}, \dots, 1 \times 10^{-4}\}$, the weight decay from $\{1 \times 10^{-2}, 5 \times 10^{-3}, \dots, 1 \times 10^{-6}\}$, the temperatures $\tau$ and $\tau_2$ from $\{5, 4, \dots, 1\}$, and the early-stopping patience from $\{50,40,\dots,10\}$. We fix the student weight decay to $0$ throughout to avoid weakening the pre-trained parameters. For simplicity, we empirically set the classification loss weight $\lambda=1$, the threshold $\eta=0.5$, the number of over-clustering $M=1.5 \times C$ in all experiments. The remaining settings are kept consistent with the selected GFM backbone. Unless otherwise specified, all experiments focus on adapting RiemannGFM to noisy node classification, while the generalization study further extends PACT to another downstream task (link prediction) and another GFM backbone (SAMGPT). For SAMGPT, LoRA is applied to the linear layers within the graph convolution modules, with $r=8$, $\alpha=16$, and a dropout rate of $0.05$. All experiments are conducted on an RTX 5090 GPU.

\subsection{Main Results on Noisy Node Classification}
\smallskip\noindent\textbf{Node classification.}
To evaluate the effectiveness and robustness of PACT for node classification under noisy supervision, we compare it with competing adaptation strategies under uniform, pair, and instance-dependent noise on four benchmark datasets. As shown in Table~\ref{tab:riemanngfm_nc_adaptation}, PACT achieves the best performance in 22 out of 24 noisy settings, demonstrating the strongest overall robustness across diverse noise patterns and intensities. The advantage is particularly evident on Cora and Citeseer, where PACT ranks first in all six settings, and becomes even more pronounced under heavier noise. For example, on Citeseer, PACT surpasses the strongest competing strategy by 8.98\% under Uniform-50\% noise and by 8.15\% under IDN-50\% noise. Moreover, on PubMed, PACT shows substantial gains under challenging noisy conditions, with the largest improvement reaching 16.54\% under IDN-50\% noise. Moreover, on the Clothing1M-graph dataset with real-world label noise, PACT still outperforms the strongest baseline (ProCon) by 1.82\%, further demonstrating its effectiveness under more complex and realistic noisy-label scenarios. It is worth noting that we construct the graph from ImageNet-pretrained ResNet-50 embeddings on the sampled subset, so the comparison mainly reflects graph learning under fixed visual representations rather than end-to-end image noisy-label learning.

These results suggest that prototype-calibrated collaborative adaptation can provide more reliable supervision under noisy labels, thereby leading to consistently stronger downstream node classification performance.

\smallskip\noindent\textbf{Few-shot results.}
To further evaluate robustness under limited supervision, we compare PACT with NRGNN- and ProCon-based adaptations under lower label budgets of 5, 10, and 15 shots per class in Table~\ref{tab:few_shot_results}. Notably, performance here is not always monotonic with respect to the label budget, since in such low-budget noisy settings, the few additional labeled samples may also introduce extra corrupted supervision, making the benefit of a larger label budget less consistent.

Overall, PACT achieves the best results in 31 out of 36 settings overall, demonstrating strong robustness when supervision is simultaneously scarce and noisy. More importantly, its improvements are often substantial rather than marginal: compared with the strongest competing baseline, PACT exceeds it by more than 5\% in 17 settings, including 9 settings with gains larger than 10\%. Even under the most challenging 5-shot regime, PACT still achieves the best performance in 10 out of 12 settings. 

\subsection{Effectiveness Analysis of PACT}
\begin{figure}[t]
\centering
\includegraphics[width=0.99\linewidth]{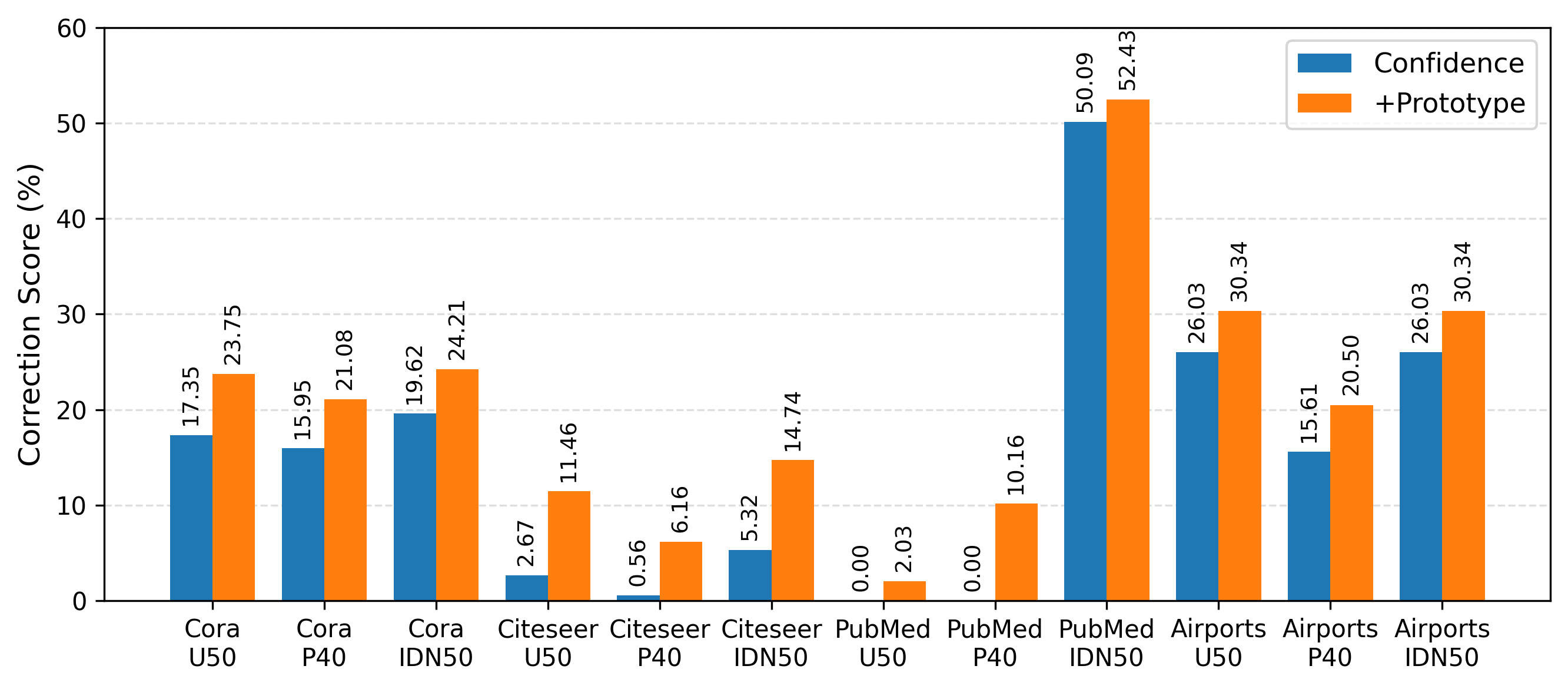}
\caption{Comparison of correction performance under different noise settings. The correction score measures how much the corrected target suppresses the original noisy label. Higher values indicate stronger correction.}
\label{fig:correction_bar}
\end{figure}

\begin{figure}[t]
\centering
\includegraphics[width=0.99\linewidth]{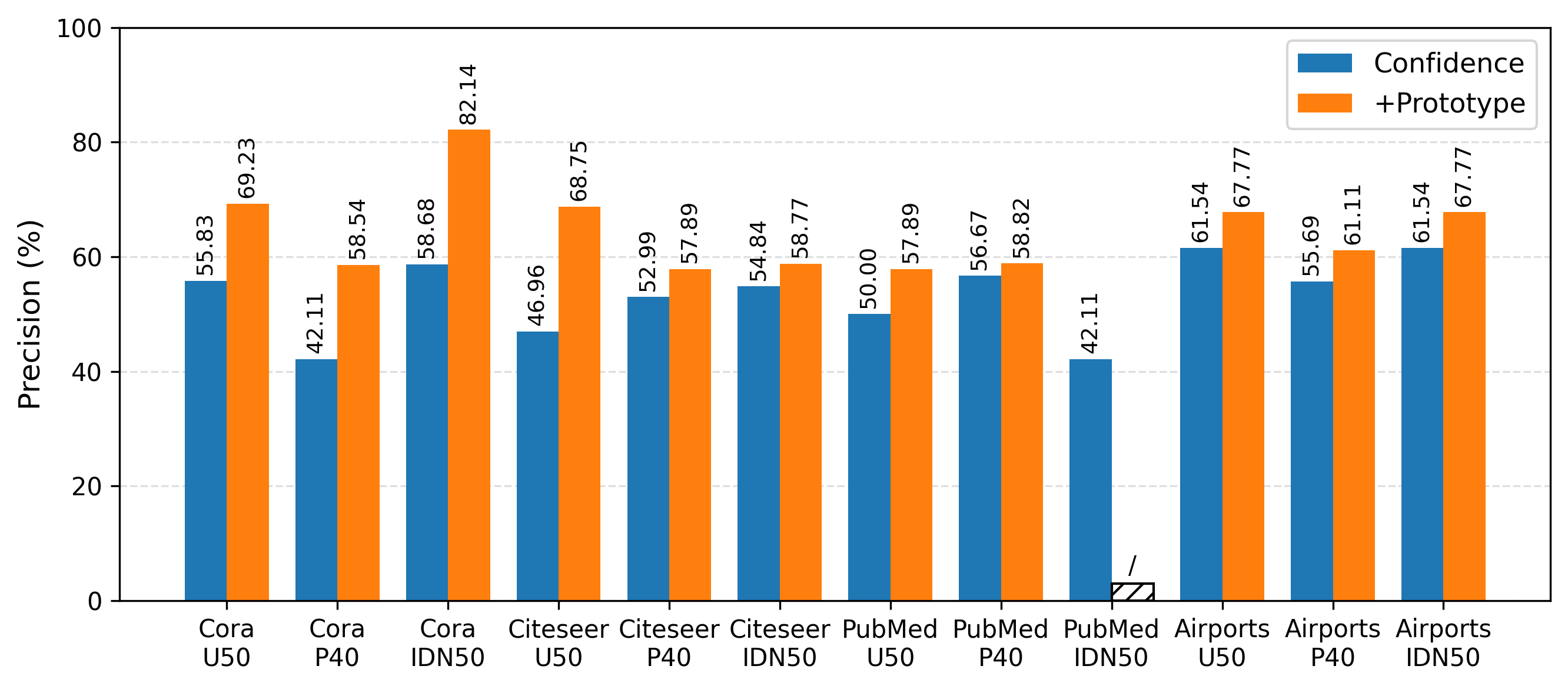}
\caption{Comparison of selection precision under different noise settings. Higher values indicate a cleaner selected subset. ``/'' denotes that no confidence sample is selected under that setting. U-*, P-*, and I-* on the x-axis denote Uniform, Pair, and IDN noise at different rates, respectively, where *(\%) indicates the corresponding noise rate.}
\label{fig:selection_bar}
\end{figure}

\smallskip\noindent\textbf{Analysis of prototype-calibrated supervision.}
To better understand how prototype calibration affects noisy adaptation, we further analyze its impact on two supervision components under high-noise settings, namely target correction and reliable sample selection. Specifically, Figure~\ref{fig:correction_bar} compares confidence-based correction with and without prototype agreement, while Figure~\ref{fig:selection_bar} compares the corresponding sample selection quality.

\begin{figure}[t]
\centering
\includegraphics[width=0.99\linewidth]{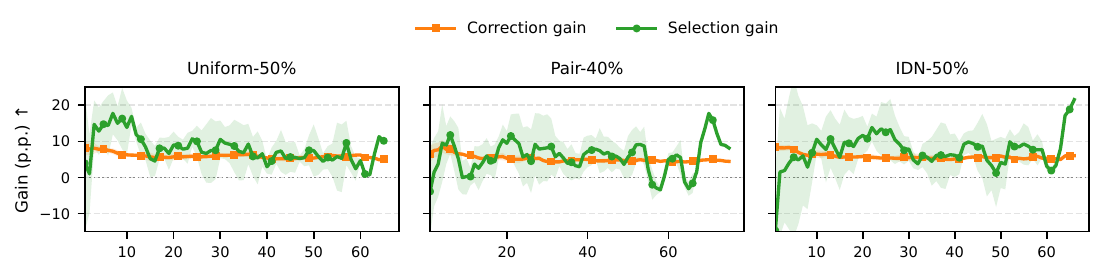}
\caption{Epoch-wise gains in correction and selection.}
\label{fig:gain_epoch}
\end{figure}

Figure~\ref{fig:correction_bar} evaluates the ability of the two strategies to suppress noisy labels, measured by the reduction in the probability assigned to the noisy label after calibration. Overall, incorporating prototype agreement leads to more effective noisy-label suppression than confidence-based correction alone. For example, at a Uniform-50\% Citeseer noise setting, the correction score improved from 2.67\% to 11.46\%, and at an IDN-50\% noise setting, the score improved from 5.32\% to 14.74\%. These results suggest that prototype-based calibration provides complementary structural cues that help weaken noisy supervision. Furthermore, as shown in Figure~\ref{fig:selection_bar}, introducing prototype guidance into confidence-based selection generally yields a cleaner selected subset across almost all settings. For instance, on the Cora dataset, selection precision improves significantly under IDN-50\% noise (from 58.68\% to 82.14\%) and shows a similar trend under Uniform-50\% noise (from 55.83\% to 69.23\%), demonstrating that prototype consistency offers a strong complementary signal for filtering unreliable samples. To further examine the training dynamics, Figure~\ref{fig:gain_epoch} reports the epoch-wise gains in correction score and selection precision brought by prototype calibration on Cora. The gains are generally positive across the three high-noise settings, showing that prototype guidance improves both supervision components throughout training. These improvements reflect the complementary roles of the two supervision components: the correction branch covers all samples and provides broad but relatively coarse noise suppression, while the selection branch focuses on a smaller subset of more reliable samples and provides cleaner hard supervision \cite{cheng2025semantic}. Together, they provide more reliable supervision for robust downstream adaptation under label noise.

\begin{figure*}[ht]
\centering
\includegraphics[width=0.95\textwidth]{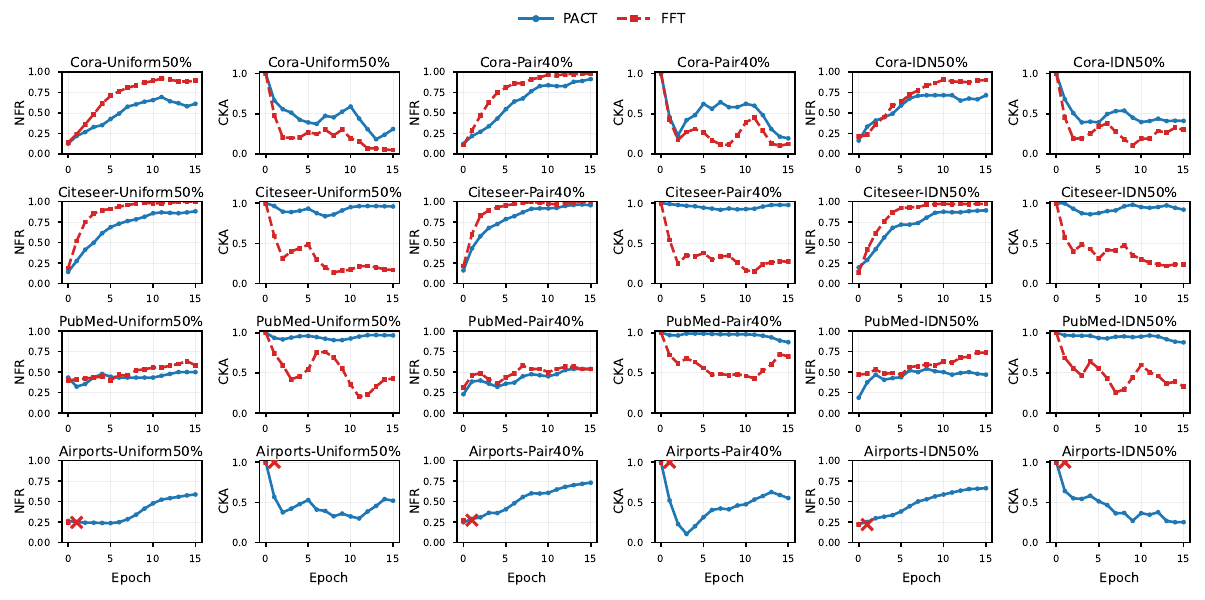}
\caption{Early-stage noise fitting and prior drift of FFT warm-up and the LoRA teacher in PACT. Here, red crosses denote non-convergent runs.}
\label{fig:earlyprobe}
\end{figure*}

\begin{table}[t]
\centering
\caption{Ablation study of PACT. The \textit{Module} group evaluates the main components of PACT, where ``w/o Cor'', ``w/o Sel'', ``w/o Pro'', ``w/o Tea'', and ``w/o Stu'' denote removing target correction, sample selection, prototype construction, the teacher branch, and the student branch, respectively. The \textit{Prototype} group further analyzes different representation sources for prototype construction, where ``w/o Raw'', ``w/o GFM'', and ``w/o T-Rep'' denote removing raw features, frozen GFM representations, and teacher representations, respectively.}
\label{tab:ablation}
\resizebox{0.99\linewidth}{!}{
\begin{tabular}{lllccc}
\toprule
Dataset & Category & Variant & Uniform-50\% & Pair-40\% & IDN-50\% \\
\midrule

\multirow{9}{*}{Cora} &  & PACT & 55.28 $\pm$ 4.50 & \textbf{47.90 $\pm$ 2.98} & \textbf{58.64 $\pm$ 2.03} \\
\cmidrule{2-6}
& \multirow{5}{*}{Module} & w/o Cor & 51.02 $\pm$ 2.39 & 44.30 $\pm$ 3.32 & 55.24 $\pm$ 1.89 \\
& & w/o Sel & 49.68 $\pm$ 2.11 & 36.60 $\pm$ 1.52 & 55.40 $\pm$ 0.83 \\
& & w/o Pro & 48.02 $\pm$ 2.82 & 32.75 $\pm$ 13.46 & 40.65 $\pm$ 14.22 \\
& & w/o Tea & 51.26 $\pm$ 1.89 & 40.84 $\pm$ 2.74 & 54.56 $\pm$ 2.43 \\
& & w/o Stu & 48.86 $\pm$ 2.00 & 44.60 $\pm$ 3.42 & 53.86 $\pm$ 1.81 \\
\cmidrule{2-6}
& \multirow{3}{*}{Prototype} & w/o Raw & 53.74 $\pm$ 1.45 & 43.92 $\pm$ 2.45 & 58.42 $\pm$ 1.59 \\
& & w/o GFM & 52.96 $\pm$ 3.85 & 47.70 $\pm$ 4.68 & 58.20 $\pm$ 1.77 \\
& & w/o T-Rep & \textbf{55.38 $\pm$ 2.80} & 46.50 $\pm$ 1.65 & 58.06 $\pm$ 0.59 \\
\midrule

\multirow{9}{*}{Citeseer} &  & PACT & \textbf{48.24 $\pm$ 2.10} & \textbf{37.62 $\pm$ 2.12} & \textbf{53.80 $\pm$ 1.65} \\
\cmidrule{2-6}
& \multirow{5}{*}{Module} & w/o Cor & 45.56 $\pm$ 5.20 & 33.86 $\pm$ 2.31 & 51.76 $\pm$ 3.03 \\
& & w/o Sel & 45.84 $\pm$ 2.19 & 33.78 $\pm$ 0.68 & 52.50 $\pm$ 1.99 \\
& & w/o Pro & 46.82 $\pm$ 4.12 & 34.40 $\pm$ 10.66 & 29.88 $\pm$ 8.23 \\
& & w/o Tea & 42.96 $\pm$ 4.01 & 30.34 $\pm$ 2.41 & 47.82 $\pm$ 2.41 \\
& & w/o Stu & 45.78 $\pm$ 2.73 & 33.52 $\pm$ 0.71 & 52.36 $\pm$ 2.53 \\
\cmidrule{2-6}
& \multirow{3}{*}{Prototype} & w/o Raw & 47.58 $\pm$ 3.40 & 36.94 $\pm$ 1.58 & 51.88 $\pm$ 2.86 \\
& & w/o GFM & 48.16 $\pm$ 4.53 & 36.66 $\pm$ 1.21 & 52.56 $\pm$ 5.16 \\
& & w/o T-Rep & 47.38 $\pm$ 2.45 & 35.54 $\pm$ 1.30 & 50.58 $\pm$ 3.59 \\
\midrule

\multirow{9}{*}{PubMed} &  & PACT & 50.66 $\pm$ 2.15 & \textbf{59.98 $\pm$ 3.78} & \textbf{46.82 $\pm$ 1.85} \\
\cmidrule{2-6}
& \multirow{5}{*}{Module} & w/o Cor & 51.72 $\pm$ 4.79 & 56.22 $\pm$ 6.75 & 42.52 $\pm$ 1.26 \\
& & w/o Sel & \textbf{54.44 $\pm$ 5.25} & 59.68 $\pm$ 1.11 & 43.78 $\pm$ 4.09 \\
& & w/o Pro & 47.58 $\pm$ 3.82 & 59.28 $\pm$ 2.04 & 39.18 $\pm$ 9.26 \\
& & w/o Tea & 45.30 $\pm$ 6.24 & 50.32 $\pm$ 4.59 & 35.46 $\pm$ 3.53 \\
& & w/o Stu & 48.60 $\pm$ 5.47 & 58.12 $\pm$ 1.93 & 46.52 $\pm$ 2.09 \\
\cmidrule{2-6}
& \multirow{3}{*}{Prototype} & w/o Raw & 52.26 $\pm$ 2.25 & 59.40 $\pm$ 1.43 & 39.78 $\pm$ 9.12 \\
& & w/o GFM & 51.62 $\pm$ 4.08 & 58.82 $\pm$ 0.78 & 44.72 $\pm$ 2.22 \\
& & w/o T-Rep & 53.08 $\pm$ 2.31 & 59.94 $\pm$ 3.17 & 46.02 $\pm$ 1.97 \\
\midrule

\multirow{9}{*}{Airports} &  & PACT & \textbf{49.16 $\pm$ 4.73} & \textbf{45.97 $\pm$ 5.00} & 46.47 $\pm$ 1.61 \\
\cmidrule{2-6}
& \multirow{5}{*}{Module} & w/o Cor & 39.75 $\pm$ 9.54 & 41.26 $\pm$ 3.67 & 37.14 $\pm$ 9.37 \\
& & w/o Sel & 47.90 $\pm$ 1.06 & 45.66 $\pm$ 3.40 & 41.68 $\pm$ 3.55 \\
& & w/o Pro & 41.51 $\pm$ 1.56 & 41.68 $\pm$ 3.55 & 36.72 $\pm$ 8.16 \\
& & w/o Tea & 39.41 $\pm$ 6.15 & 43.19 $\pm$ 3.50 & 38.66 $\pm$ 3.17 \\
& & w/o Stu & 44.29 $\pm$ 3.24 & 41.34 $\pm$ 3.12 & 44.29 $\pm$ 3.12 \\
\cmidrule{2-6}
& \multirow{3}{*}{Prototype} & w/o Raw & 48.50 $\pm$ 3.36 & 45.72 $\pm$ 2.28 & 45.63 $\pm$ 4.09 \\
& & w/o GFM & 48.25 $\pm$ 3.90 & 45.97 $\pm$ 6.39 & \textbf{49.41 $\pm$ 1.52} \\
& & w/o T-Rep & 46.39 $\pm$ 1.52 & 44.71 $\pm$ 4.64 & 44.37 $\pm$ 4.24 \\
\bottomrule
\end{tabular}
}
\end{table}

\begin{figure}[t]
\centering
\includegraphics[width=0.4\linewidth]{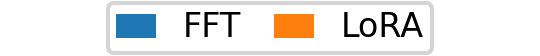}

\subfloat[]{
    \includegraphics[width=0.22\linewidth]{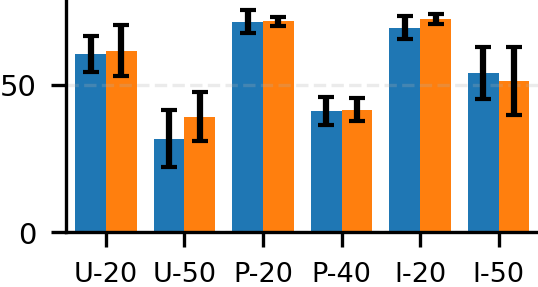}
}
\subfloat[]{
    \includegraphics[width=0.22\linewidth]{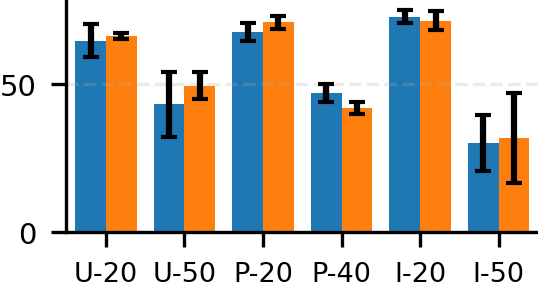}
}
\subfloat[]{
    \includegraphics[width=0.22\linewidth]{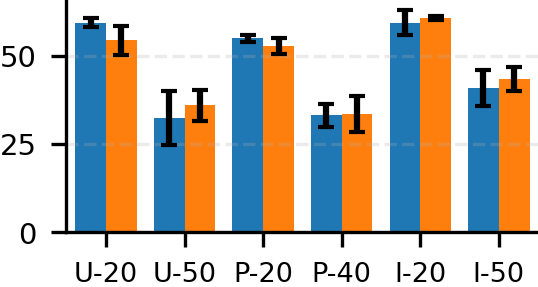}
}
\subfloat[]{
    \includegraphics[width=0.22\linewidth]{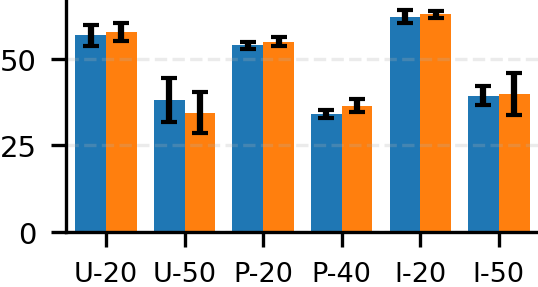}
}
\hfill
\subfloat[]{
    \includegraphics[width=0.22\linewidth]{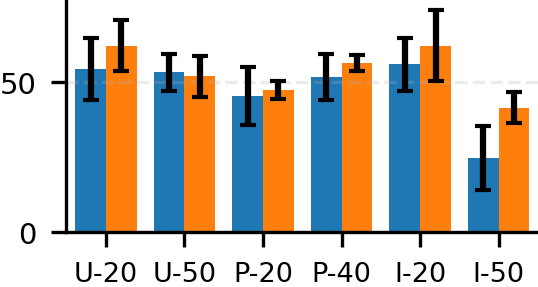}
}
\subfloat[]{
    \includegraphics[width=0.22\linewidth]{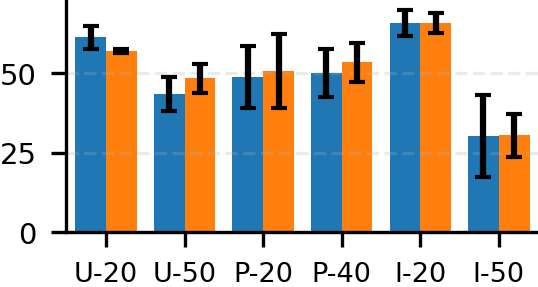}
}
\subfloat[]{
    \includegraphics[width=0.22\linewidth]{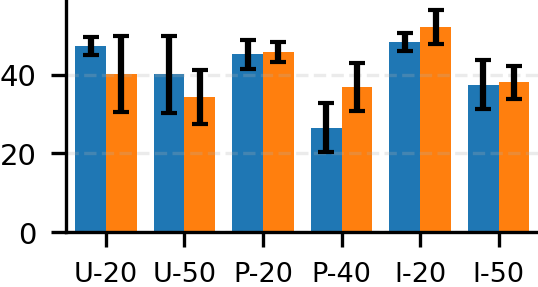}
}
\subfloat[]{
    \includegraphics[width=0.22\linewidth]{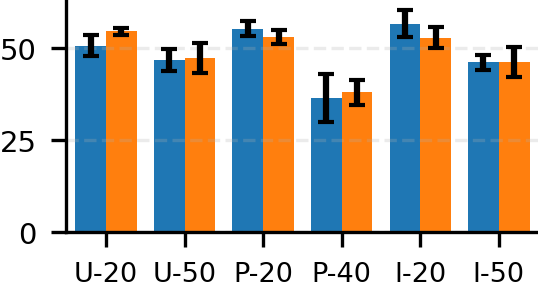}
}
\hfill
\caption{Comparison of FFT and LoRA with different noisy-label learning strategies across datasets and noise settings: (a) Cora-NRGNN; (b) Cora-ProCon; (c) CiteSeer-NRGNN; (d) CiteSeer-ProCon; (e) PubMed-NRGNN; (f) PubMed-ProCon; (g) Airports-NRGNN; and (h) Airports-ProCon. Here, U-*, P-*, and I-* on the x-axis denote Uniform, Pair, and IDN noise, respectively, where * denotes the noise rate (\%).}
\label{fig:lora_fft_comparison}
\end{figure}

\smallskip\noindent\textbf{Early-stage Warm-up Analysis.}
To examine whether PACT alleviates early damage to pre-trained priors caused by premature noise fitting, we further compare FFT warm-up with the LoRA teacher in PACT during the early adaptation stage under three high-noise settings. Specifically, in Fig.~\ref{fig:earlyprobe}, we report the noise fitting rate, $\mathrm{NFR}_t=\frac{1}{|\mathcal{N}|}\sum_{i\in\mathcal{N}} \mathbb{I}[\hat{y}_i^{(t)}=\tilde{y}_i]$, where $\mathcal{N}$ is the set of corrupted training nodes, $\tilde{y}_i$ is the observed noisy label, and $\hat{y}_i^{(t)}$ is the prediction at epoch $t$; we also compute the linear CKA similarity between the centered geometric representations of the frozen GFM, $\widetilde{Z}^{(0)}$, and those at epoch $t$, $\widetilde{Z}^{(t)}$, as $\mathrm{CKA}_t=\frac{
\left\|\widetilde{Z}^{(0)\top}\widetilde{Z}^{(t)}\right\|_F^2}{\left\|\widetilde{Z}^{(0)\top}\widetilde{Z}^{(0)}\right\|_F\left\|\widetilde{Z}^{(t)\top}\widetilde{Z}^{(t)}\right\|_F}$.

As shown in Fig.~\ref{fig:earlyprobe}, FFT warm-up rapidly fits corrupted labels in the early epochs, and this process is accompanied by a clear decrease in CKA similarity. This indicates that direct FFT warm-up can quickly deviate from the pre-trained representation structure while fitting noisy supervision. In contrast, the LoRA teacher in PACT fits fewer corrupted labels and maintains higher CKA similarity to the frozen GFM during early adaptation. These results directly support our motivation that PEFT-based teacher adaptation avoids premature noisy-label fitting and better protects useful pre-trained priors before reliable supervision is constructed.

\smallskip\noindent\textbf{Ablation study.}
To investigate the roles of the individual components of PACT, we present ablation results under three representative strong-noise settings in Table~\ref{tab:ablation}. The module ablations show that both prototype-aware target correction and agreement-aware sample selection generally improve performance, while removing either the teacher or the student also leads to clear degradation, demonstrating the effectiveness of the collaborative teacher-student design. Among these variants, removing the prototype module causes the most significant degradation in many settings, especially under the more challenging IDN noise, verifying that prototype construction provides critical class-level guidance for noisy adaptation. The prototype ablations further show that all three representation sources generally contribute to prototype quality, while their relative benefits vary across settings. This is because they provide complementary information from different perspectives: raw features preserve the original data-level information, frozen GFM representations capture transferable knowledge learned from the source domain, and teacher representations provide task-adaptive information from the target domain. In particular, the former two are isolated from downstream noisy-label optimization, while the latter is learned through the more stable teacher branch. Their combination therefore integrates data-, source-domain-, and target-domain information to construct more reliable prototypes under label noise.

We note that a few ablated variants slightly outperform the full model in specific settings, such as removing sample selection on PubMed under Uniform-50\% noise and removing the frozen GFM representation on Airports under IDN-50\% noise. These results suggest that the optimal strength of sample filtering and the relative utility of different prototype sources can vary across settings. In practice, overly restrictive sample selection may discard still-informative instances and reduce supervision coverage, while an additional representation source may provide redundant or less task-relevant information when its contribution overlaps with those of other sources. This helps explain why removing a specific component or representation source can occasionally be beneficial in individual settings. Although further validation-based tuning could better balance these contributions for specific settings, we use the complete model with a unified configuration throughout for simplicity and consistency, as it provides more robust overall performance across diverse datasets and noise types.

\smallskip\noindent\textbf{Analysis of adaptation regimes.}
To further analyze the robustness of PEFT and FFT for noisy GFM adaptation under representative noisy-label learning guidance, we compare LoRA and FFT within NRGNN and ProCon. As shown in Figure~\ref{fig:lora_fft_comparison}, replacing FFT with LoRA generally yields stronger performance, with LoRA outperforming FFT in 34 out of 48 settings. This suggests that PEFT is overall a more robust adaptation regime under noisy supervision. 

Nevertheless, the improvement remains relatively limited, suggesting that simply substituting the adaptation regime is still insufficient for noisy GFM adaptation, even when combined with state-of-the-art noisy-label learning strategies. In particular, such plug-and-play combinations still fail to simultaneously exploit the stronger adaptation capacity of FFT and the better robustness of PEFT. This further motivates the need for a dedicated collaborative design that jointly performs noise filtering and supervision calibration.

\begin{table*}[t]
\centering
\caption{Results on noisy link prediction under uniform and instance-dependent noise. The best result is highlighted in bold.}
\label{tab:lp_results}
\resizebox{0.95\textwidth}{!}{
\begin{tabular}{llcccccccc}
\toprule
\multirow{2}{*}{Dataset} & \multirow{2}{*}{Method}
& \multicolumn{2}{c}{Uniform-20\%}
& \multicolumn{2}{c}{Uniform-50\%}
& \multicolumn{2}{c}{IDN-20\%}
& \multicolumn{2}{c}{IDN-50\%} \\
\cmidrule(lr){3-4}\cmidrule(lr){5-6}\cmidrule(lr){7-8}\cmidrule(lr){9-10}
& & ROC-AUC & PR-AUC & ROC-AUC & PR-AUC & ROC-AUC & PR-AUC & ROC-AUC & PR-AUC \\
\midrule
\multirow{4}{*}{Cora}
& LoRA   & 76.29 ± 1.06 & 83.01 ± 0.44 & 62.27 ± 0.56 & 71.51 ± 0.77 & 80.93 ± 0.61 & 85.64 ± 0.23 & 62.32 ± 1.94 & 69.83 ± 1.21 \\
& FFT    & 76.03 ± 0.65 & 82.85 ± 0.34 & 62.23 ± 1.14 & 70.95 ± 1.33 & 81.06 ± 0.64 & 85.67 ± 0.64 & 62.38 ± 0.79 & 69.55 ± 1.53 \\
& ProCon & 70.61 ± 0.72 & 72.00 ± 0.95 & \textbf{68.64 ± 1.67} & 70.42 ± 2.21 & 71.94 ± 1.78 & 72.96 ± 2.30 & \textbf{70.33 ± 1.20} & 70.93 ± 1.40 \\
& PACT   & \textbf{77.97 ± 0.40} & \textbf{83.18 ± 0.73} & 63.80 ± 1.08 & \textbf{71.90 ± 0.42} & \textbf{81.40 ± 1.52} & \textbf{85.88 ± 1.32} & 63.12 ± 0.99 & \textbf{71.66 ± 0.89} \\
\midrule
\multirow{4}{*}{Citeseer}
& LoRA   & 80.92 ± 1.06 & 87.40 ± 0.93 & 63.73 ± 1.57 & 74.41 ± 0.87 & 84.21 ± 0.61 & 89.26 ± 0.67 & 61.76 ± 0.40 & 73.39 ± 0.31 \\
& FFT    & 80.91 ± 0.94 & 87.41 ± 0.76 & 64.97 ± 1.46 & 75.32 ± 1.28 & 85.43 ± 0.51 & 90.18 ± 0.38 & 63.68 ± 1.09 & 74.91 ± 0.46 \\
& ProCon & 75.74 ± 0.38 & 79.29 ± 0.56 & \textbf{71.84 ± 0.73} & 75.09 ± 0.55 & 75.41 ± 1.63 & 78.70 ± 1.58 & \textbf{71.52 ± 1.20} & 74.75 ± 0.75 \\
& PACT   & \textbf{83.55 ± 0.67} & \textbf{87.72 ± 0.84} & 67.20 ± 1.69 & \textbf{75.59 ± 1.27} & \textbf{88.28 ± 1.13} & \textbf{91.41 ± 1.38} & 65.86 ± 2.49 & \textbf{75.01 ± 0.95} \\
\midrule
\multirow{4}{*}{PubMed}
& LoRA   & 84.02 ± 1.68 & 90.30 ± 0.37 & 57.17 ± 3.21 & 73.19 ± 0.85 & 88.00 ± 2.60 & 91.71 ± 0.97 & 58.08 ± 5.11 & 70.98 ± 1.08 \\
& FFT    & 82.22 ± 0.79 & 90.00 ± 0.15 & 55.74 ± 2.49 & 72.51 ± 0.13 & 85.65 ± 2.06 & 90.99 ± 0.58 & 55.42 ± 3.22 & 68.87 ± 1.37 \\
& ProCon & 67.56 ± 10.04 & 68.52 ± 9.32 & 61.17 ± 1.58 & 62.93 ± 1.10 & 61.72 ± 2.01 & 63.46 ± 1.10 & 62.53 ± 3.16 & 64.05 ± 2.52 \\
& PACT   & \textbf{87.52 ± 0.70} & \textbf{90.51 ± 0.50} & \textbf{72.43 ± 10.34} & \textbf{78.97 ± 8.18} & \textbf{90.39 ± 0.63} & \textbf{92.46 ± 0.45} & \textbf{75.71 ± 8.88} & \textbf{82.03 ± 6.74} \\
\midrule
\multirow{4}{*}{Airports}
& LoRA   & 55.48 ± 5.75 & 64.41 ± 8.92 & 36.06 ± 3.95 & 54.19 ± 4.95 & 62.35 ± 5.05 & 74.34 ± 6.75 & 45.69 ± 1.34 & 64.88 ± 2.80 \\
& FFT    & 64.31 ± 4.39 & 75.30 ± 8.92 & 46.33 ± 2.02 & 65.01 ± 3.56 & 58.49 ± 3.22 & 65.44 ± 6.94 & 35.93 ± 2.41 & 52.93 ± 5.45 \\
& ProCon & 43.56 ± 1.03 & 46.29 ± 0.90 & 39.69 ± 2.36 & 44.00 ± 1.13 & 41.56 ± 2.58 & 45.48 ± 1.76 & 39.42 ± 2.39 & 43.92 ± 1.57 \\
& PACT   & \textbf{91.24 ± 6.39} & \textbf{94.72 ± 3.55} & \textbf{68.65 ± 21.22} & \textbf{80.55 ± 13.21} & \textbf{89.34 ± 5.23} & \textbf{92.67 ± 4.24} & \textbf{85.08 ± 12.51} & \textbf{90.34 ± 8.03} \\
\bottomrule
\end{tabular}
}
\end{table*}

\subsection{Generalization and Transferability}
\smallskip\noindent\textbf{Generalizing to link prediction task.} In this section, we adapt PACT to noisy link prediction to examine its generalizability across downstream tasks and compare it with representative adaptation baselines under uniform and instance-dependent noise settings. We formulate each candidate link as a binary classification instance, where observed edges are treated as positive links and an equal number of negative links are sampled from unconnected node pairs. The training and validation link labels are corrupted under the same noise setting, while the test labels remain clean. Since link prediction is a binary task, uniform and pair noise are effectively equivalent. For IDN, the edge-level noise feature is constructed by concatenating the raw features of the two endpoint nodes. The link predictor first projects the node representations and uses the cosine similarity between the projected endpoint representations as the link score. For prototype construction, we concatenate the prototype representations of the two endpoint nodes to form the representation of each link and apply the same prototype calibration and collaborative supervision procedure as in node classification. All models are trained for up to $200$ epochs with a default batch size of $64$ and a learning rate of $1\times10^{-3}$, and the best checkpoint is selected according to validation PR-AUC. Here, we do not include NRGNN in this comparison, because its core self-supervised link reconstruction introduces label leakage, while disabling link reconstruction reduces it to a standard GCN.

As shown in Table~\ref{tab:lp_results}, PACT exhibits strong transferability to link prediction and achieves the best overall performance across datasets and noise settings. Specifically, it obtains the best PR-AUC in all 16 evaluated cases, and also achieves the best ROC-AUC in 12 of them. The gains are particularly significant on PubMed and Airports, where PACT substantially outperforms all baselines under higher noise levels. This shows that, even when the supervision signal shifts from node labels to edge labels, the collaborative tuning strategy and prototype-based calibration in PACT remain effective. Although ProCon achieves higher ROC-AUC on Cora and Citeseer under extremely severe noise, PACT attains the best PR-AUC in these cases, indicating better retrieval quality for positive links under noisy supervision. These results suggest that the advantage of PACT is not limited to a specific downstream task.

\begin{table*}[t]
\centering
\caption{Results of adapting SAMGPT under different label noise settings. The best results are in \textbf{bold}.}
\label{tab:samgpt_generalization}
\begin{tabular}{llcccccc}
\toprule
Dataset & Method & Uniform-20\% & Uniform-50\% & Pair-20\% & Pair-40\% & IDN-20\% & IDN-50\% \\
\midrule
\multirow{3}{*}{Cora}
& NRGNN  & \textbf{75.48 ± 1.67} & 50.84 ± 1.71 & \textbf{76.40 ± 1.89} & 44.78 ± 4.06 & 73.18 ± 1.33 & 54.08 ± 1.45 \\
& ProCon & 66.44 ± 3.14 & 41.46 ± 6.13 & 72.54 ± 0.86 & 44.20 ± 2.79 & \textbf{76.74 ± 1.31} & 45.56 ± 6.06 \\
& PACT   & 70.82 ± 1.36 & \textbf{52.36 ± 0.93} & 72.32 ± 1.01 & \textbf{45.86 ± 3.34} & 73.68 ± 1.50 & \textbf{62.06 ± 1.82} \\
\midrule
\multirow{3}{*}{Citeseer}
& NRGNN  & 65.44 ± 1.50 & 46.92 ± 0.20 & 55.44 ± 0.97 & 36.54 ± 0.29 & 65.34 ± 1.92 & 46.36 ± 0.24 \\
& ProCon & 64.90 ± 1.64 & 34.76 ± 8.05 & 53.80 ± 1.08 & 36.12 ± 1.66 & 63.68 ± 1.49 & 45.00 ± 4.35 \\
& PACT   & \textbf{65.50 ± 2.16} & \textbf{47.06 ± 3.50} & \textbf{56.10 ± 2.71} & \textbf{37.64 ± 3.33} & \textbf{65.58 ± 1.64} & \textbf{47.06 ± 3.50} \\
\midrule
\multirow{3}{*}{PubMed}
& NRGNN  & 57.24 ± 0.42 & 42.34 ± 2.82 & 71.44 ± 3.40 & 63.00 ± 1.36 & 67.08 ± 0.87 & 17.06 ± 0.31 \\
& ProCon & 57.78 ± 0.53 & 40.30 ± 0.88 & 60.90 ± 11.97 & \textbf{64.22 ± 0.55} & 67.12 ± 1.55 & 20.14 ± 4.90 \\
& PACT   & \textbf{58.28 ± 1.46} & \textbf{48.72 ± 4.18} & \textbf{71.64 ± 1.74} & 58.28 ± 4.71 & \textbf{67.66 ± 1.50} & \textbf{41.38 ± 3.92} \\
\midrule
\multirow{3}{*}{Airports}
& NRGNN  & 49.66 ± 1.87 & 43.78 ± 1.34 & 48.66 ± 2.21 & 44.62 ± 2.34 & 50.25 ± 0.87 & 21.43 ± 0.84 \\
& ProCon & 53.28 ± 1.34 & 44.45 ± 7.40 & 53.05 ± 1.69 & 43.36 ± 3.58 & 51.76 ± 1.47 & 36.72 ± 9.20 \\
& PACT   & \textbf{54.45 ± 2.47} & \textbf{48.66 ± 3.65} & \textbf{53.36 ± 1.38} & \textbf{47.73 ± 2.97} & \textbf{55.63 ± 0.98} & \textbf{42.27 ± 2.39} \\
\bottomrule
\end{tabular}
\end{table*}

\smallskip\noindent\textbf{Generalizing to other GFM.} 
To further evaluate backbone-level generalization, we additionally adapt all methods to SAMGPT and compare them with recent noisy-label learning strategies under the same adaptation setting. As shown in Table~\ref{tab:samgpt_generalization}, our method still achieves the best performance in 20 out of 24 settings. Notably, its advantage becomes particularly clear under severe noise. In the Uniform-50\% and IDN-50\% settings, our method improves over the strongest baseline by an average of 3.06\% and 8.87\%, respectively. This trend indicates that the proposed collaborative adaptation and prototype-based correction mechanisms remain effective even when transferred to a structurally different GFM. Meanwhile, on several relatively easier settings on Cora, such as Uniform-20\%, Pair-20\%, and IDN-20\%, NRGNN or ProCon still achieves slightly better results. This further suggests that the main strength of our method does not lie in stronger fitting under mild noise, but in better robustness when noisy supervision becomes more destructive. Overall, these results provide strong evidence that PACT generalizes beyond a single backbone and can serve as a broadly applicable adaptation framework for noisy-label learning with GFMs.

\begin{figure}[t]
\centering
\includegraphics[width=0.7\linewidth]{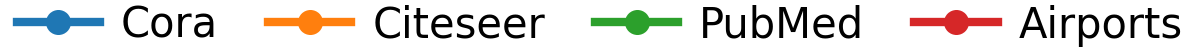}
\hfill
\subfloat[$\eta$]{
    \includegraphics[width=0.22\linewidth]{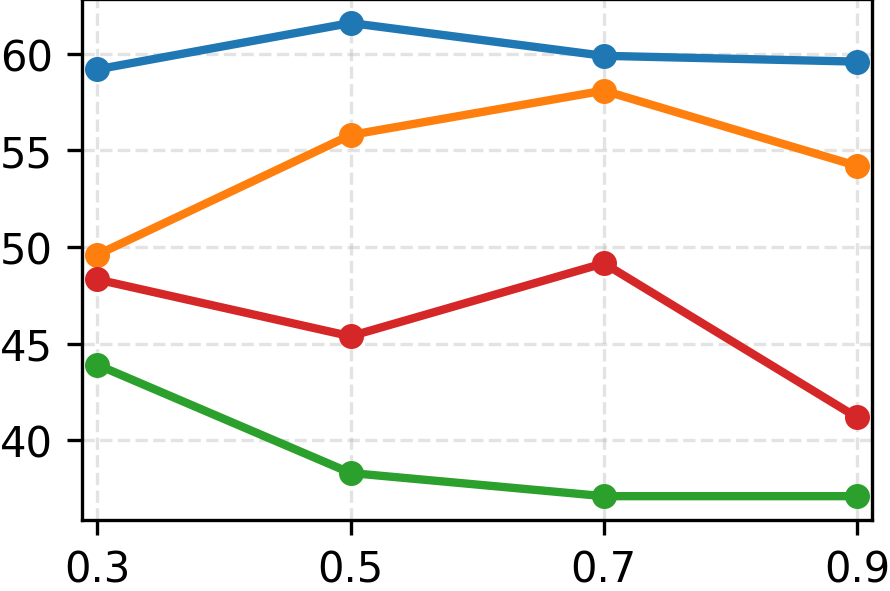}
} 
\subfloat[$\lambda$]{
    \includegraphics[width=0.22\linewidth]{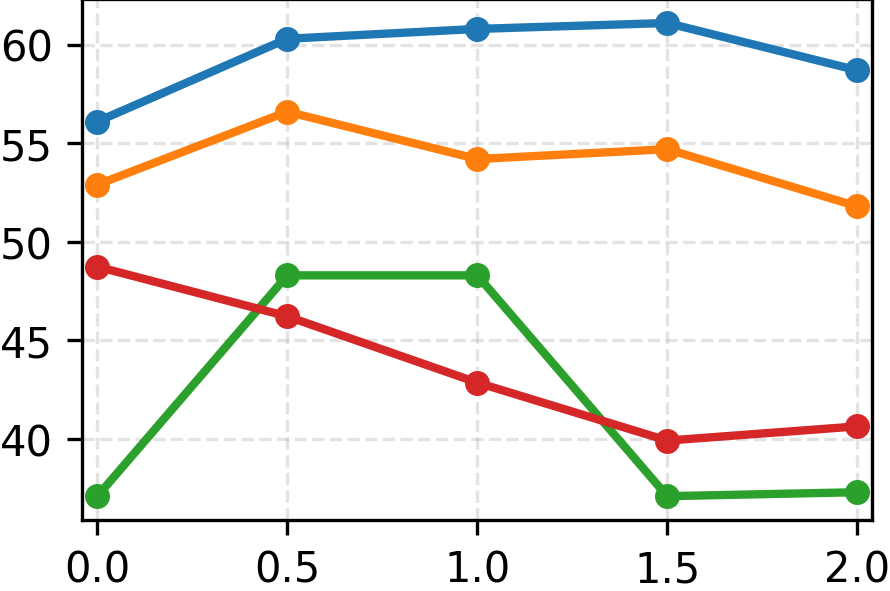}
}
\subfloat[$M^r$]{
    \includegraphics[width=0.22\linewidth]{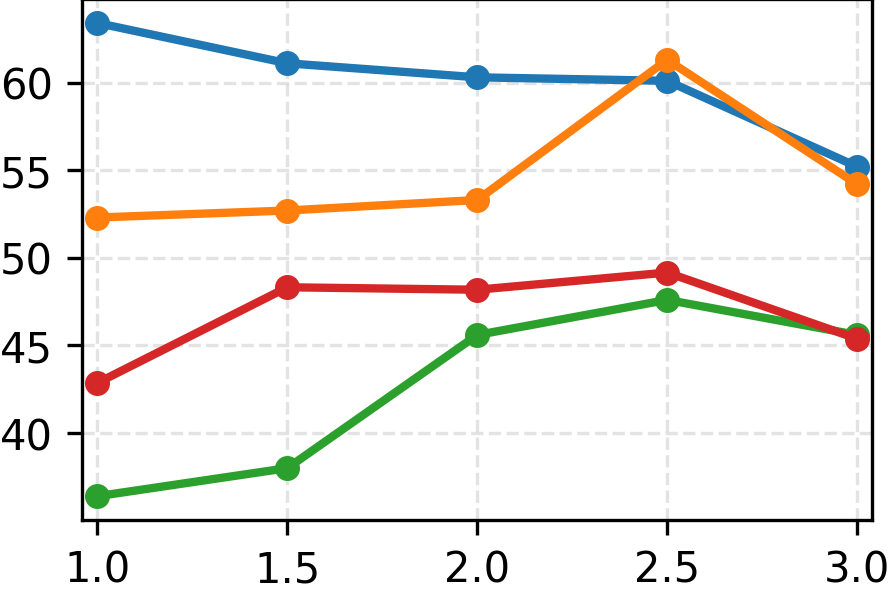}
}
\subfloat[$\tau$]{
    \includegraphics[width=0.22\linewidth]{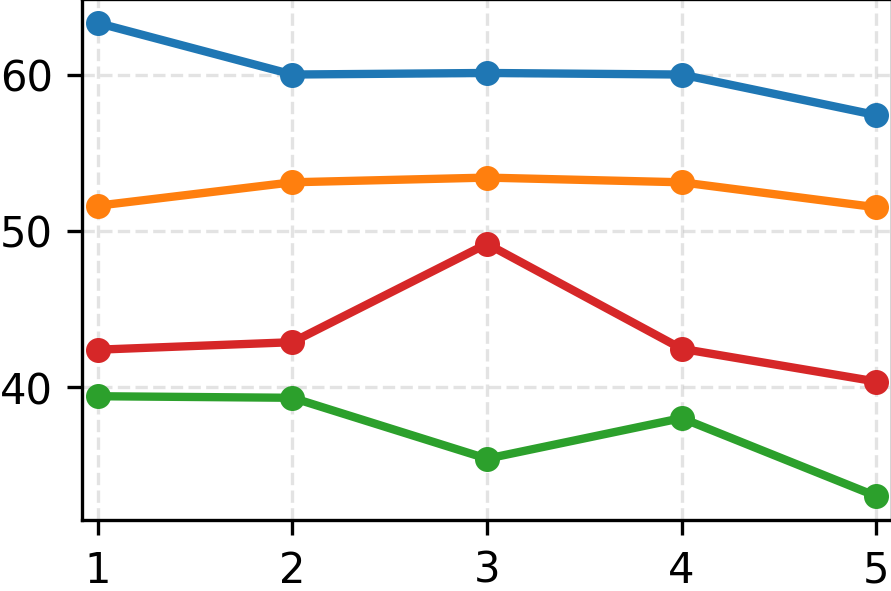}
}
\caption{Parameter sensitivity of PACT on four datasets.}
\label{fig:sensitive}
\end{figure}

\subsection{Parameter Sensitivity Analysis}
We further analyze the sensitivity of PACT to the over-cluster rate $M^r$ (with the number of clusters set to $M=M^r \times C$), the temperature $\tau$ (with $\tau_2=\tau$ for simplicity), the reliability threshold $\eta$, and the weight $\lambda$ for the loss on selected samples. As shown in Figure~\ref{fig:sensitive}, PACT remains relatively stable across a reasonably wide range of values on all four datasets, indicating good robustness to hyperparameter choices.

\section{Conclusion}
In this paper, we investigated the problem of adapting Graph Foundation Models under label noise. To prevent unreliable downstream supervision from prematurely corrupting transferable pre-trained knowledge, we proposed Prototype-calibrated Adaptation with Collaborative Tuning (PACT), which couples a robust LoRA teacher with a high-capacity FFT student through prototype-guided supervision calibration. Extensive experiments show that PACT consistently improves noisy adaptation and generalizes across supervision budgets, downstream tasks, and GFM backbones. More importantly, the gains are generally larger when reliable supervision for each class becomes sparse or difficult to distinguish, as can occur under severe noise, larger label spaces, instance-dependent corruption, or limited supervision. In such cases, existing methods have greater difficulty forming stable class-level references from noisy downstream labels alone, making prototype-guided calibration more beneficial. In contrast, the improvements are relatively modest when sufficient coherent supervision remains for each class and reliable decision boundaries can already be learned. These observations also highlight a limitation of PACT: its prototype calibration benefits from transferable task-relevant information retained in the frozen GFM representations. When the pre-trained GFM itself has limited transferability to a substantially different downstream task or domain, such information may be less sufficient to support prototype construction, leaving room for further improvement through adaptive integration of different representation sources. In future work, we will investigate such adaptive integration, as well as joint adaptation over multiple noisy source graphs, to improve robust zero- or few-shot generalization to unseen target graphs.

\section*{Acknowledgments}
This research was partially supported by the Key Research and Development Project in Shaanxi Province No. 2024PTZCK-89, the National Science Foundation of China No. 62406242, 62672397 and 62476215.

\bibliographystyle{IEEEtran}
\bibliography{ref}

\end{document}